\documentclass[manuscript,sigconf,nonacm]{acmart}
\usepackage[ruled,vlined]{algorithm2e}
\usepackage{subcaption}
\usepackage{float}
\AtBeginDocument{%
  }

\begin{document}

\title{Addressing and Visualizing Misalignments in Human Task-Solving Trajectories}


\author{Sejin Kim}
\email{sejinkim@gist.ac.kr}
\affiliation{%
  \institution{GIST}
  \city{Gwangju}
  \country{Korea}
}

\author{Hosung Lee}
\email{confeitohs@gmail.com}
\affiliation{%
  \institution{KAIST}
  \city{Daejeon}
  \country{Korea}
}

\author{Sundong Kim}
\authornote{Corresponding author.}
\email{sundong@gist.ac.kr}
\affiliation{%
  \institution{GIST}
  \city{Gwangju}
  \country{Korea}
}

\renewcommand{\shortauthors}{Kim et al.}

\begin{abstract}
Understanding misalignments in human task-solving trajectories is crucial for enhancing AI models trained to closely mimic human reasoning.
This study categorizes such misalignments into three types: \textbf{(1) lack of functions to express intent}, \textbf{(2) inefficient action sequences}, and \textbf{(3) incorrect intentions that cannot solve the task}.  
To address these issues, we first formalize and define these three misalignment types in a unified framework.  
We then propose a heuristic algorithm to detect misalignments in ARCTraj trajectories and analyze their impact hierarchically and quantitatively.
We also present an intention estimation method based on our formalism that infers missing alignment between user actions and intentions.
Through trajectory alignment, we experimentally demonstrate that AI models trained on human task-solving trajectories improve performance in mimicking human reasoning.  
Based on hierarchical analysis and experiments, we highlight the importance of trajectory-intention alignment and demonstrate the effectiveness of intention-aligned training.
\end{abstract}



\maketitle

\newcommand\kddavailabilityurlA{https://doi.org/10.5281/zenodo.15515271}
\newcommand\kddavailabilityurlB{https://doi.org/10.5281/zenodo.15515295}

\section*{Available Links} 
The experiment code is available at \url{\kddavailabilityurlA}.
The data analysis code is available at \url{\kddavailabilityurlB}.

\section*{Additional Resources}
To support research on human-like reasoning in ARC tasks, we present \textbf{ARCTraj}, a curated dataset of human task-solving trajectories~\cite{kim2025arctraj}. Collected via structured web interfaces, ARCTraj captures diverse strategies, temporal dependencies, and behavioral variability observed during problem solving. The dataset is freely available on Hugging Face at \url{https://huggingface.co/datasets/SejinKimm/ARCTraj}, along with an interactive viewer at \url{https://arc-traj-viewer.vercel.app/}, which enables intuitive exploration of trajectory patterns and user strategies. For details on dataset construction and its use in evaluating human-like reasoning, see the ARCTraj paper~\cite{kim2025arctraj}, available at \url{https://arc-traj-viewer.vercel.app/ARCTraj_paper.pdf}.

\section{Introduction}
Developing AI models capable of human-like reasoning is a fundamental goal in the field of artificial intelligence research. These systems should perform tasks efficiently and adapt to novel situations with the flexibility and generalization that are characteristic of human cognition. ARC-AGI has emerged as a pivotal benchmark for assessing these abilities in AI models~\cite{chollet2019ARC}.

\begin{figure}[htbp!]
    \centering
    \includegraphics[width=0.99\columnwidth]{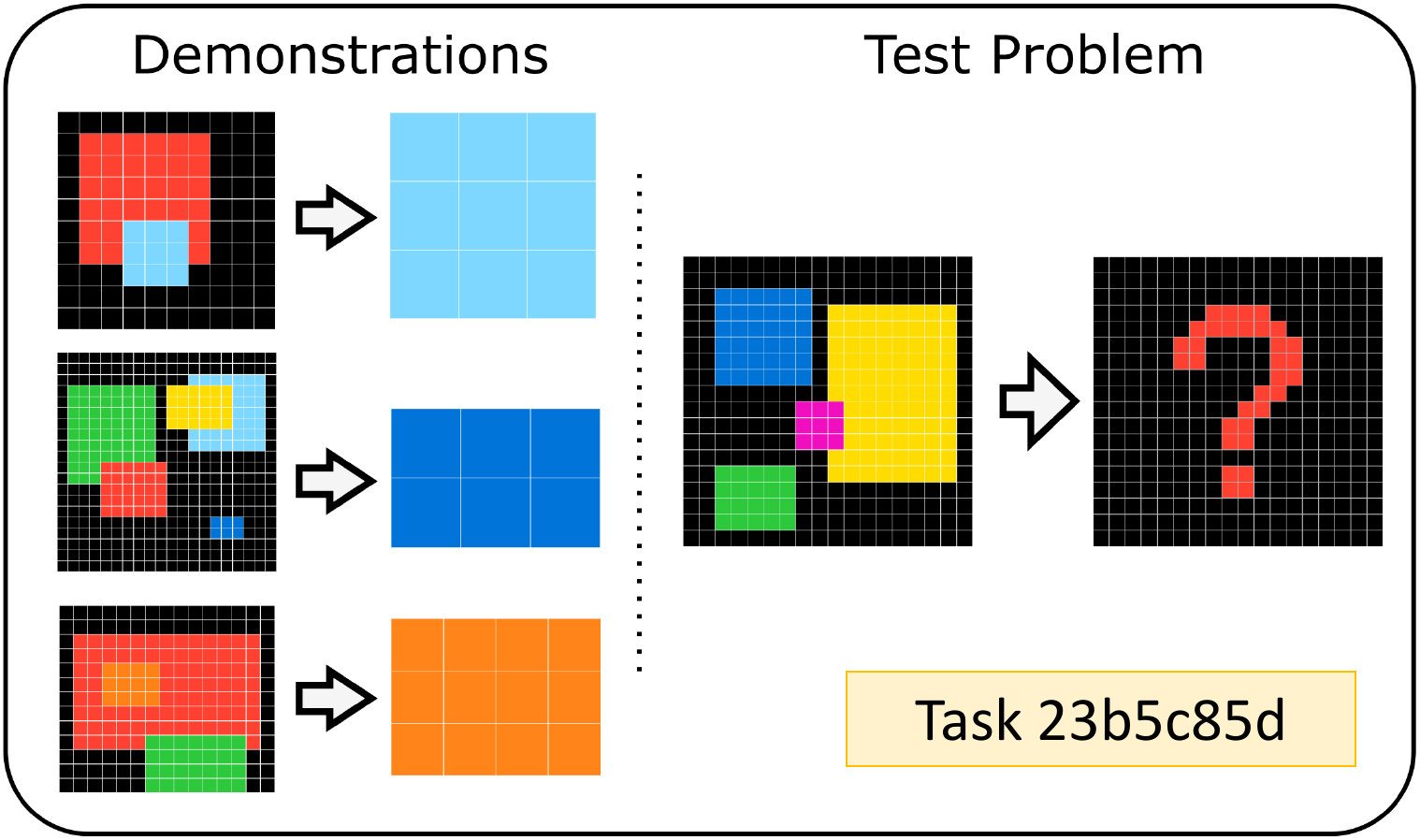}
    \caption{An example of ARC task (Task 23b5c85d). The goal is to infer the rule from input-output pairs and generate the correct output grid for a new input. The correct answer is the $3 \times 3$ magenta-colored rectangle.}
    \Description{An example of ARC task (Task 23b5c85d). The goal is to infer the rule from input-output pairs and generate the correct output grid for a new input. The correct answer is the $3 \times 3$ magenta-colored rectangle.}
    \label{fig:example_arc_task}
\end{figure}

As shown in Fig.~\ref{fig:example_arc_task}, ARC tasks require identifying high-level transformations from minimal examples and applying them to new inputs, closely mirroring human cognitive processes. Despite significant advancements in deep learning and reinforcement learning, current models face challenges in effectively solving ARC tasks, highlighting a persistent gap between AI and human reasoning.

Recent large language models (LLMs) have demonstrated impressive performance on the ARC benchmark. However, their approach remains highly inefficient due to substantial computational costs per problem. For instance, Falcon-40B achieved 61.9\% accuracy among publicly available LLMs, while OpenAI o3 reached 87.5\%. These results come at an enormous computational expense, requiring orders of magnitude more computation than human problem-solving, underscoring the need for AI models that can perform reasoning more efficiently and in a human-like manner.

To address this, interactive interfaces~\cite{borsky2021arcgame, johnson2021fast, lab422022arcreate, kim2022playgrounds, shim2024o2arc, neoneye2024arc-interactive, legris2024harc} collect human task-solving trajectories, offering insights into human approaches. 
Among them, the ARCTraj dataset~\cite{kim2025arctraj}, collected via the O2ARC platform~\cite{shim2024o2arc}, provides extensive human trajectory data and forms the basis for ARCLE~\cite{lee2024arcle}, a reinforcement learning environment designed to model human strategies.

However, training AI models directly on ARCTraj trajectories presents challenges. Recent studies~\cite{park2023unraveling, kim2024diffusion} utilize these trajectories, but their effectiveness is limited by rigid preprocessing rules, reducing generalization to unseen tasks. We hypothesize that this stems from \textbf{misalignments between human intentions and trajectory data}, caused by tool limitations, reasoning-to-action gaps, and even user errors. We categorize these misalignments into three types, each corresponding to a distinct challenge in human task-solving, as framed by Activity Theory~\cite{leontiev1978activity}:

\begin{itemize}
    \item \textbf{Functional Inadequacies in Tools} corresponds to the \textit{lack of functions to express intent}. In many cases, users intend to execute specific actions, but the tools provided do not allow them to express these intentions directly.  As a result, they must resort to workarounds or indirect methods, leading to unintended inefficiencies in their trajectories.

    \item \textbf{User Unfamiliarity with Tools} aligns with the \textit{choice of inefficient action sequences}. Due to limited familiarity with the toolset, users may not always choose the most optimal sequence of actions. Instead, they experiment or take unnecessary steps, resulting in redundant or overly complex trajectories.

    \item \textbf{Cognitive Dissonance in Users}: relates to \textit{incorrect intentions that cannot solve the task}. Users may misinterpret task objectives or apply incorrect strategies, leading them to take actions that ultimately fail to solve the problem. Unlike the previous categories, these errors stem from conceptual misunderstandings rather than tool constraints.
\end{itemize}

To systematically address trajectory misalignments, we first formalize three types of misalignment. Specifically, we introduce popular states and ideal actions to characterize inefficiencies in human task-solving trajectories. Using this framework, we analyze misalignment patterns in ARCTraj, identifying their prevalence across multiple levels—action, intention, trajectory, and task.

To mitigate these misalignments, we propose an \textbf{Intention Prediction Algorithm} (Alg.~\ref{alg:intention_prediction}) that aligns trajectories with inferred human intentions. By detecting key \textit{popular states} and encoding their transitions, we infer structured intention labels for each action.

Furthermore, we empirically validate that intention-aligned supervision enhances the efficiency of AI learning. Models trained with inferred intention labels exhibit improved generalization, reducing reliance on spurious correlations and promoting structured decision-making. This study aligns AI learning with human problem-solving strategies, establishing a framework for trajectory-based learning and highlighting the broader implications of \textbf{Intention Learning} for AI models seeking to achieve human-like reasoning.  

\paragraph{Contributions} This study makes the following key contributions to trajectory-based AI learning. 
\begin{itemize} \item We formalize and define three primary types of misalignment in human task-solving trajectories. \item We conduct a hierarchical and quantitative analysis of misalignment patterns. \item We propose an Intention Prediction Algorithm to align trajectories with inferred human intentions. \item We empirically demonstrate that intention-aligned trajectory learning improves AI task-solving performance. \end{itemize}

\section{Related Work}

\subsection{ARC Tasks and Human Trajectories}
The Abstraction and Reasoning Corpus for Artificial General Intelligence (ARC-AGI)~\cite{chollet2019ARC} serves as a benchmark for evaluating AI models' abstract reasoning and generalization capabilities. While earlier AI models, including deep learning-based approaches, achieved at most 55--60\% accuracy on ARC tasks, recent advancements in large language models (LLMs) have pushed performance to 87.5\% using test-time reasoning techniques~\cite{akyurek2024tttarc}. However, these models process a large number of tokens at inference time, making their approach fundamentally different from human reasoning and significantly less cost-efficient. This discrepancy highlights the need for AI systems that perform efficient, human-like reasoning rather than relying on brute-force search over extensive token sequences.

Human task-solving trajectories offer valuable insights for AI learning~\cite{johnson2021fast, legris2024harc}. 
The ARCTraj dataset~\cite{kim2025arctraj}, collected via the O2ARC platform~\cite{shim2024o2arc}, provides extensive human trajectories that have enabled reinforcement learning environments such as ARCLE~\cite{lee2024arcle} to model human strategies.  
To date, ARCTraj is one of the largest available corpora capturing fine-grained user actions on ARC tasks, covering various strategies and behaviors. Its scale and diversity make it particularly suitable for analyzing patterns of reasoning and error.

However, previous studies~\cite{park2023unraveling, lee2024analogical, kim2024diffusion} faced scalability and generalization issues due to rigid preprocessing rules and single-task limitations.  
Language-based approaches~\cite{acquaviva2022communicating, moskvichev2023conceptARC} explore structured guidance through explanations or concept grouping, but they lack direct mappings to user actions.  
These efforts offer high-level reasoning cues but often overlook the procedural structure inherent in human trajectories.

This study addresses these limitations by systematically analyzing and mitigating misalignments in the ARCTraj dataset.

\subsection{Activity Theory and Misalignment Analysis}
Misalignment between user intentions and actions is a key challenge in AI training~\cite{norman1995psychopathology}. 
We leverage Activity Theory~\cite{leontiev1978activity} to systematically analyze these discrepancies.

Activity Theory frames human activity as interactions between the user (subject), task (object), and tools (instruments), with misalignments arising from contradictions in this triad:
\begin{itemize}
    \item \textbf{Functional Inadequacies in Tools}: When tools lack necessary functions, users resort to workarounds.
    \item \textbf{User Unfamiliarity with Tools}: Limited proficiency leads to inefficient action sequences.
    \item \textbf{Cognitive Dissonance in Users}: Misunderstanding task objectives results in incorrect actions.
\end{itemize}

This study focuses on the direct impact of these misalignments on reasoning and task-solving performance, excluding broader sociocultural components of Activity Theory.
By identifying these recurring patterns, we aim to improve alignment between user intent and model behavior during trajectory-based learning.
Our approach helps bridge the gap between human demonstrations and structured AI learning, ensuring that models capture user intentions rather than merely replicating surface-level behaviors.
\section{Formalizing Misalignments in Human Trajectories}

The ARCTraj dataset~\cite{kim2025arctraj} provides detailed records of users' decisions and actions as they solve ARC tasks.
Each trajectory is structured using the state-action sequences defined by the ARC Learning Environment (ARCLE)~\cite{lee2024arcle}, which serves as a framework for analyzing user interactions and problem-solving strategies.

\subsection{Format of Human Trajectories}
To formalize ARC tasks and trajectories, we adopt the notation proposed in recent work~\cite{akyurek2024tttarc}. 
A single ARC task $d$ from the set of tasks $\mathcal{D}_{\text{ARC}}$ is defined as:
\begin{equation*}
    d = \{ (\mathbf{x}_k^{\text{train}}, \mathbf{y}_k^{\text{train}})_{k=1}^K, (\mathbf{x}^{\text{test}}, \mathbf{y}^{\text{test}}) \} \in \mathcal{D}_{\text{ARC}},
\end{equation*}
where $(\mathbf{x}_k^{\text{train}}, \mathbf{y}_k^{\text{train}})$ are input-output grid pairs provided as demonstrations, and $(\mathbf{x}^{\text{test}}, \mathbf{y}^{\text{test}})$ represent the test problem grid and its answer grid.

\begin{figure}[htbp!]
    \centering
    \includegraphics[width=\columnwidth]{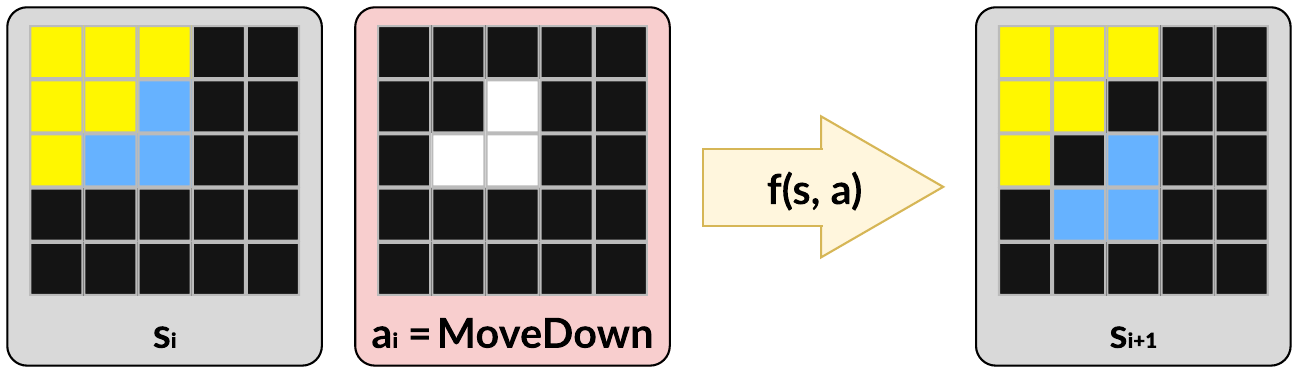}
    \caption{A single state transition step in ARCLE~\cite{lee2024arcle}. An action transforms the current state $s_i$ into the next state $s_{i+1}$ through the transition function $f(s_i, a_i) = s_{i+1}$. In this example, the selected grids masked in white are shifted down by one row.}
    \Description{A single state transition step in ARCLE~\cite{lee2024arcle}. An action transforms the current state $s_i$ into the next state $s_{i+1}$ through the transition function $f(s_i, a_i) = s_{i+1}$. In this example, the selected grids masked in white are shifted down by one row.}
    \label{fig:arcle_state_transition}
\end{figure}

User trajectories $\mathcal{T}_d$ are collected as state-action sequences while solving a task $d$. 
A single trajectory $\tau_d$ is represented as:
\begin{equation*}
    \tau_d = (s_0, a_0, s_1, a_1, \dots, s_n) \in \mathcal{T}_d,
    \label{eq:trajectory}
\end{equation*}
where $s_i$ denotes a state, $a_i$ an action, and $n$ the number of actions in the trajectory $\tau_d$.
Every state $s_i$ and action $a_i$ in the trajectory $\tau_d$ follow the same state-action transition function $f(s, a)$ as the yellow arrow in Fig.~\ref{fig:arcle_state_transition}:
\begin{equation*}
    f(s_i, a_i) = s_{i+1}
\end{equation*}

\paragraph{States} Each state $s_i$ is derived from $\mathbf{x}^{\text{test}}$ and evolves through the state-action transition function $f(s, a)$ like:
\begin{equation*}
    s_i = \begin{cases}
        \mathbf{x}^{\text{test}} & \text{if } i = 0, \\
        f(s_{i-1}, a_{i-1}) & \text{if } 1 \leq i \leq n.
    \end{cases}
    \label{eq:state}
\end{equation*}
States encompass the current task grid and additional contextual information, such as object properties and clipboard contents, which enable actions like copying and pasting.
These additional signals provide richer data for analyzing user strategies~\cite{lee2024arcle}.

This formalization captures the dynamic nature of human problem-solving. As discussed in subsequent sections, it serves as a conceptual foundation for identifying potential misalignments between user actions and inferred intentions.

\subsection{Fundamental Concepts}
The human trajectory does not explicitly record user intentions, making it harder to analyze and quantify misalignments between actions and intentions.
We address this by inferring intentions from trajectories using structural patterns and key assumptions.

We assume that users generating each trajectory $\tau_d$ pursue similar problem-solving strategies.
Although reasoning diversity exists, ARC tasks emphasize precise analogies, often leading users to converge on shared subgoals.
While this may oversimplify, prior studies~\cite{johnson2021fast, legris2024harc} observed consistent patterns across tasks.

Some actions may belong to multiple intentions, but such overlaps occur in a few trajectories and have minimal impact.
Likewise, some misaligned sub-trajectories may reflect creative or efficient alternatives, offering rare yet valuable insights for future work.

\paragraph{Popular States}
\textit{Popular states}, also called bottleneck states in prior studies~\cite{johnson2021fast}, are defined as frequently visited states among trajectories.
We identify popular states using a threshold $\theta(|\mathcal{T}_d|)$ based on the number of trajectories for task $d$.
For example, a common choice like $\theta(x) = \sqrt{x}$~\footnote{
$\sqrt{x}$ is a widely used heuristic for filtering frequent items under Zipfian distributions. In NLP, it serves as a vocabulary cutoff; in Information Retrieval, it guides keyword pruning and index reduction; and in graph or web log analysis, it selects salient patterns. Our ablation results show that varying $\theta(x)$ among $\log x$, $\sqrt{x}$, and $0.2x$ yields similar numbers of popular states (11.4, 6.5, and 7.1, respectively), suggesting robustness to this choice.
} strikes a balance between sensitivity and robustness in identifying frequently visited states.
\begin{equation*}
    \mathcal{P}_d = \{ s_i \mid N(s_i; \mathcal{T}_d) \geq \theta(|\mathcal{T}_d|) \},
\end{equation*}
where $N(s_i; \mathcal{T}_d)$ denotes the number of trajectories visiting state $s_i$.

\paragraph{Intention}
An \textit{intention} is defined as a sequence of actions that transitions between two popular states.
An action sequence $a_{i:j}$ consists of consecutive actions $a_i, a_{i+1}, \dots, a_{j-1}$ that transition from $s_i$ to $s_j$ while avoiding intermediate popular states. Formally:
\begin{align*}
    &a_{i:j} = (a_i, a_{i+1}, \dots, a_{j-1}) \text{ such that } \nonumber \\
    &s_i, s_j \in \mathcal{P}_d \text{ and } s_k \notin \mathcal{P}_d \text{ for } i < k < j.
\end{align*}

\paragraph{Ideal Actions}
To simplify the representation of the intention, we define an \textit{ideal action}, denoted as $a_{i:j}^\star$. 
This ideal action encapsulates the intention behind the sequence $a_{i:j}$, abstracting the sequence into a single, efficient action that transitions directly from $s_i$ to $s_j$.
\begin{equation*}
    f(s_i, a_{i:j}^\star) = s_j.
\end{equation*}
Here, $a_{i:j}^\star$ is a hypothetical action summarizing the intention of $a_{i:j}$. 
The set of all such ideal actions for $d$ is:
\begin{equation*}
    \mathcal{A}_d^\star = \{ a_{i:j}^\star \mid f(s_i, a_{i:j}^\star) = s_j \text{ where } s_i, s_j \in \mathcal{P}_d \}.
\end{equation*}

\paragraph{Ideal Trajectory}
An \textit{ideal trajectory} is defined as a sequence of popular states and ideal actions that transition through these states, satisfying the conditions for an optimal transition. 
Unlike observed trajectories in $\mathcal{T}_d$, ideal trajectories represent a conceptual framework for understanding optimal user behavior. 
Thus, the ideal trajectory $\tau_d^\star$ for a task $d$ is represented as:
\begin{equation*}
    \tau_d^\star = (s_0, a_0, s_1, a_1, \dots, s_n) \text{ where } \forall s_i \in \mathcal{P}_d, \forall a_i \in \mathcal{A}_d^\star.
\end{equation*}
Here, $\tau_d^\star$ represents the optimal path that fully aligns with the user's intentions as captured by ideal actions.

\begin{figure*}[htbp!]
    \centering
    \subfloat[Ideal Trajectory]{
        \includegraphics[width=0.48\textwidth]{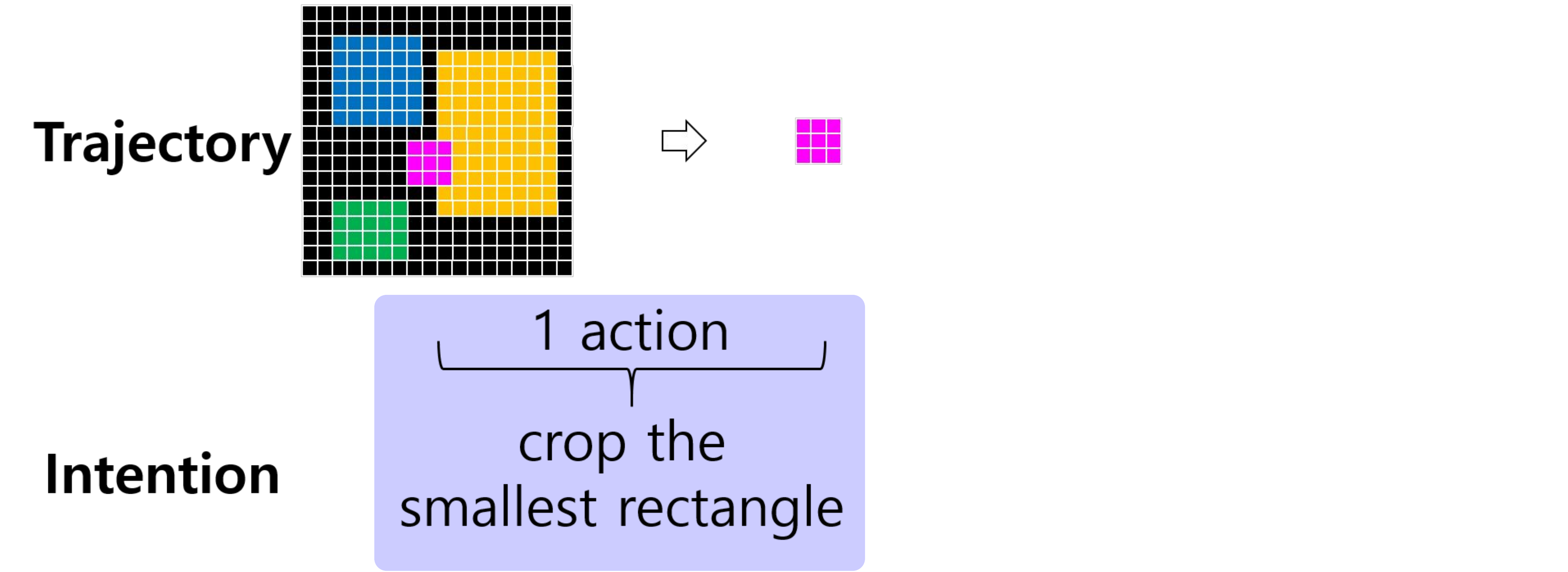}
        \label{fig:best_answer}
    }
    \subfloat[Functional Inadequacies in Tools]{
        \includegraphics[width=0.48\textwidth]{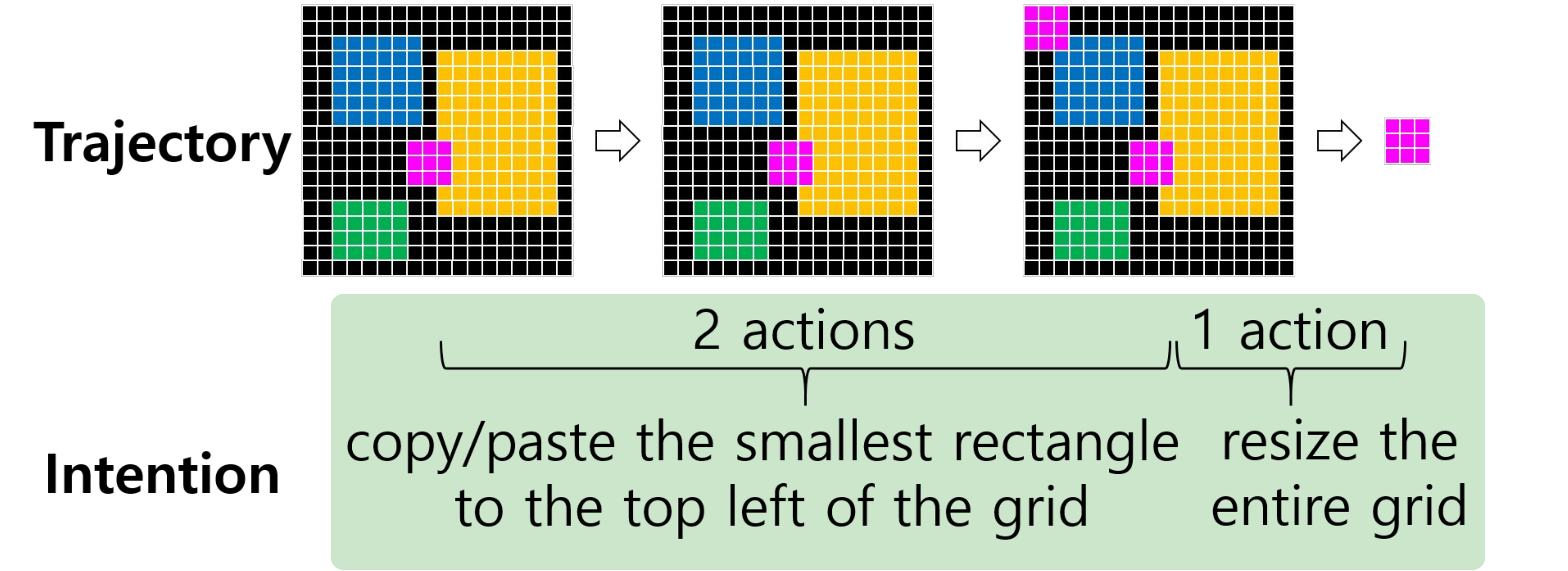}
        \label{fig:tool_limit}
    }
    \\
    \subfloat[User Unfamiliarity with Tools]{
        \includegraphics[width=0.48\textwidth]{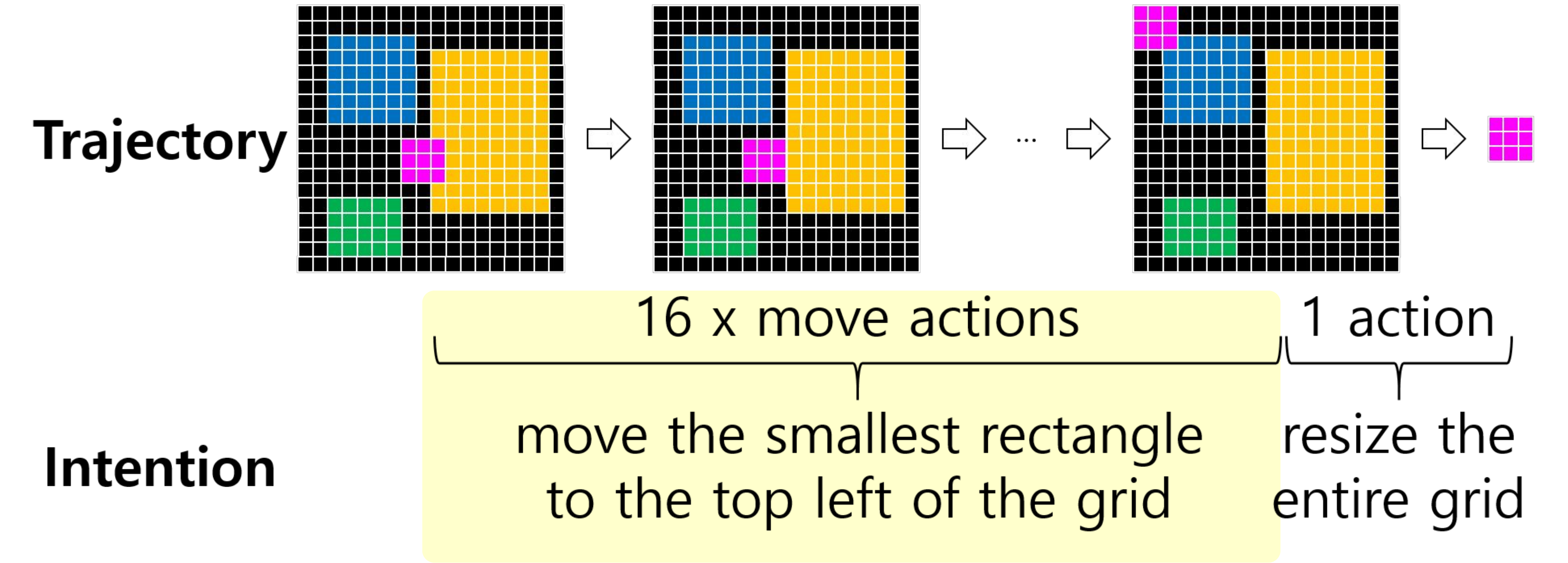}
        \label{fig:user_limit}
    }
    \subfloat[Cognitive Dissonance in Users]{
        \includegraphics[width=0.48\textwidth]{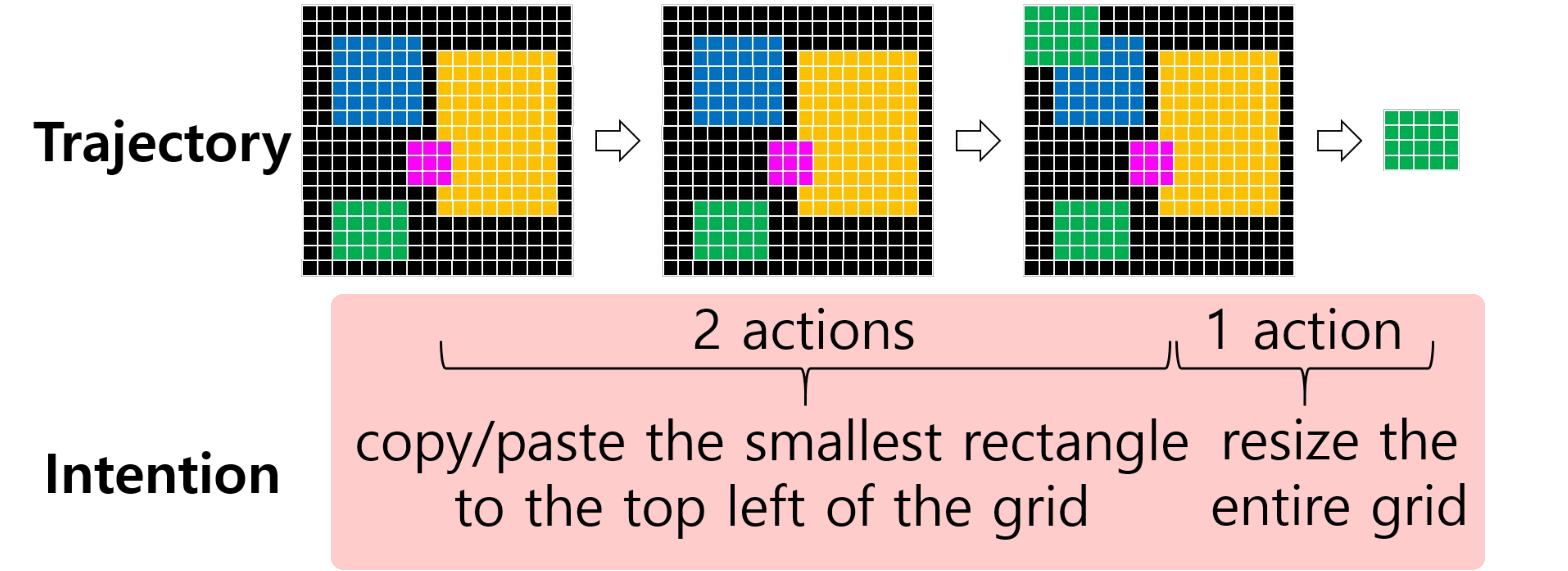}
        \label{fig:user_wrong}
    }
    \caption{Various trajectories for ARC Task 23b5c85d as shown in Fig.~\ref{fig:example_arc_task}. (a) The ideal trajectory transitions directly between popular states with the shortest possible sequence of actions, perfectly representing user intentions. (b) Functional inadequacies influence the trajectory in tools, where the lack of a suitable action requires combining multiple actions, resulting in longer transitions between popular states. (c) A trajectory is caused by user unfamiliarity with tools, where redundant actions reflect inefficiencies despite a shorter ideal trajectory. (d) Trajectory reflects cognitive dissonance in users, where errors or misinterpretations prevent reaching the correct answer state, deviating from transitions between popular states.}
    \Description{Various trajectories for ARC Task 23b5c85d as shown in Fig.~\ref{fig:example_arc_task}. (a) The ideal trajectory transitions directly between popular states with the shortest possible sequence of actions, perfectly representing user intentions. (b) Functional inadequacies influence the trajectory in tools, where the lack of a suitable action requires combining multiple actions, resulting in longer transitions between popular states. (c) A trajectory is caused by user unfamiliarity with tools, where redundant actions reflect inefficiencies despite a shorter ideal trajectory. (d) Trajectory reflects cognitive dissonance in users, where errors or misinterpretations prevent reaching the correct answer state, deviating from transitions between popular states.}
    \label{fig:trajectory_types}
\end{figure*}

\subsection{Three Types of Misalignments}
Based on Activity Theory~\cite{engestrom2015learning}, these misalignments can be categorized into three distinct types, each reflecting contradictions between users, tools, and tasks in the ARC task-solving process.
We formalize and analyze these categories using the concepts of popular states, inferred intentions, ideal actions, and ideal trajectories introduced in the previous section.

\paragraph{Functional Inadequacies in Tools}
This misalignment stems from tool–task mismatch.
The toolset (e.g., O2ARC) lacks an ideal action $a_{i:j}^\star$ linking popular states $s_i$ and $s_j$.
\begin{equation*}
    a_{i:j}^\star \notin \mathcal{A}_{d}.
\end{equation*}
As a result, users must combine a sequence of actions $(a_i, \dots, a_{j-1})$ to achieve the same intention:
\begin{equation*}
    a_{i:j} = (a_i, a_{i+1}, \dots, a_{j-1}), \text{ where } a_k \in \mathcal{A}_d \setminus \mathcal{A}_d^\star \; \text{for } i \leq k < j.
\end{equation*}
For instance, Fig.~\ref{fig:tool_limit} shows a trajectory where the absence of a ``cropping'' action forces the user to rely on sequences such as copying, pasting, and resizing. 
This results in an inefficient trajectory compared to the ideal trajectory shown in Fig.~\ref{fig:best_answer}, where direct transitions are achieved using ``cropping''.

\paragraph{User Unfamiliarity with Tools}
This misalignment arises from contradictions between users and tools. 
Although an ideal action $a_{i:j}^\star$ exists in the supported action set $\mathcal{A}_{d}$:
\begin{equation*}
    a_{i:j}^\star \in \mathcal{A}_{d},
\end{equation*}
Then, the user skips it and takes an inefficient action sequence.
\begin{equation*}
    a_{i:j} = (a_i, a_{i+1}, \dots, a_{j-1}) \text{ where } a_k \in \mathcal{A}_d \setminus \mathcal{A}_d^\star \; \text{for } i \leq k < j.
\end{equation*}
Suppose the trajectory in Fig.~\ref{fig:tool_limit} represents an ideal trajectory. Fig.~\ref{fig:user_limit} then illustrates a scenario where the user, unfamiliar with the copy-paste functionality, attempts to move an object step by step. This results in a redundant action sequence of 16 actions, which could otherwise be represented by two ideal actions, highlighting inefficiency in transitioning between popular states.

\paragraph{Cognitive Dissonance in Users}
This misalignment reflects contradictions between users and tasks. 
Cognitive Dissonance can manifest in two distinct cases: (1) when users have incorrect intentions or (2) when users possess the correct intention but execute incorrect actions, leading them to an incorrect final state. 
However, since human trajectories do not explicitly record user intentions, verifying the second case is not feasible. 
Therefore, we define and detect Cognitive Dissonance based on the first case.

Specifically, Cognitive Dissonance occurs when a user’s trajectory fails to reach a correct final state, deviating from popular states:
\begin{equation*}
s_n \neq \mathbf{y}^{\text{test}}, \; \text{ where } \mathbf{y}^{\text{test}} \in \mathcal{P}_d
\end{equation*}
Here, $s_n$ represents the final state of the trajectory, $\mathbf{y}^{\text{test}}$ the answer grid of the task $d$, and $\mathcal{P}_d$ the set of popular states for the task $d$.  
\textit{This typically holds since correct solutions tend to be convergent and often visited by multiple users.}

This focus highlights cases where the user misunderstood the task objectives or made significant errors during problem-solving, resulting in an incorrect solution state.
Such deviations not only hinder success but also reflect deeper issues in perception, strategy, or tool use.
Analyzing these cases can reveal patterns of confusion or bias, offering insights for improving instruction, interfaces, or even curricula.

\begin{figure}[htbp!]
\centering
\includegraphics[width=0.68\columnwidth]{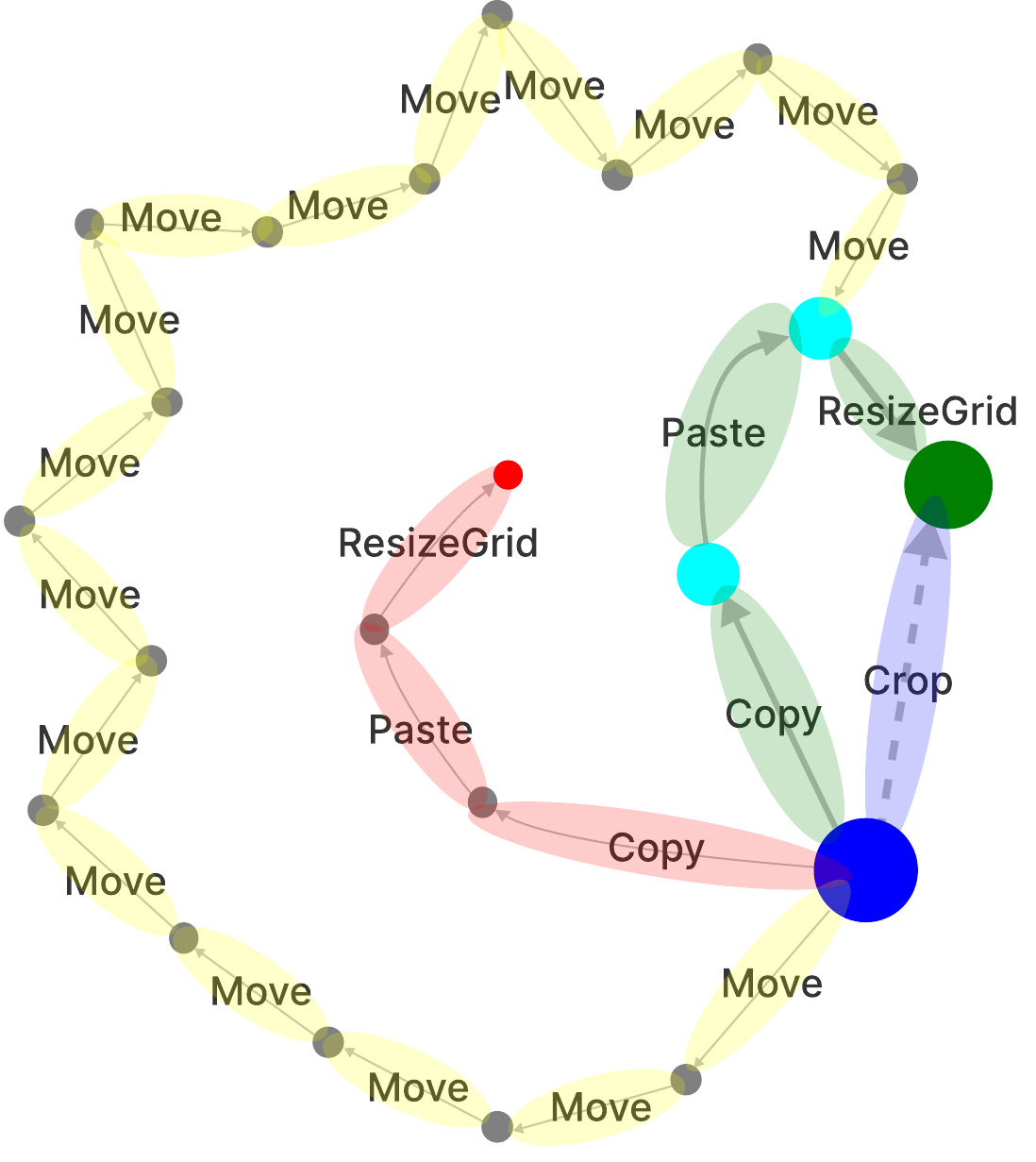}
\caption{The state space graph of user trajectories from Fig.~\ref{fig:trajectory_types}. The blue node represents the test problem state, the green node denotes the answer state, and the red nodes indicate states where incorrect answers were submitted. The thickness of nodes and edges reflects the frequency of occurrence for each respective state and action.}
\Description{The state space graph of user trajectories from Fig.~\ref{fig:trajectory_types}. The blue node represents the test problem state, the green node denotes the answer state, and the red nodes indicate states where incorrect answers were submitted. The thickness of nodes and edges reflects the frequency of occurrence for each respective state and action.}
\label{fig:trajectory_visualization}
\end{figure}

\subsection{State Space Graph}
Fig.~\ref{fig:trajectory_visualization} visualizes user trajectories from Fig.~\ref{fig:trajectory_types} in a state space graph. 
Each node represents a distinct state encountered during task solving, and directed edges represent the actions taken between these states. 
The size of each node reflects the frequency of its visits, while the thickness of each edge indicates the frequency of the corresponding action sequence. 
This provides a concise yet informative overview of the distribution and transitions in human behavior.

Edges with the same color across Fig.~\ref{fig:best_answer}--\ref{fig:user_wrong} are blurred to emphasize deviation from the ideal trajectory.
Misalignment types are color-coded: blue for Functional Inadequacies, yellow for User Unfamiliarity, and red for Cognitive Dissonance.
This scheme facilitates the intuitive identification of error types across trajectories and enables a comparative inspection of how different reasoning errors are distributed throughout the task-solving process.

The classification of some edges depends on whether intermediate cyan nodes are considered popular states. 
Suppose only the blue and green nodes are included in $\mathcal{P}_d$. In that case, the dashed blue edge represents a missing ideal action ($\xrightarrow{crop}$), making the green and yellow paths examples of Functional Inadequacies in Tools. 
These paths require multiple actions to achieve what could be done by a single ideal action. 
This illustrates how certain inefficiencies stem not from user intent but from limitations in available action primitives that constrain expressivity.

Alternatively, if cyan nodes are also considered popular, each green edge becomes an ideal action. 
The yellow edge ($\xrightarrow{move} \times 16$) now reflects User Unfamiliarity with Tools—inefficiency despite better available actions. 
Such cases reveal how misalignments stem from both system design and user behavior. 
They help identify opportunities for training or interface improvements and distinguish between structural and correctable inefficiencies.
%This dual perspective underscores the importance of modeling assumptions in defining and interpreting user intentions.

\paragraph{Visualization with State Space Graph} 
To illustrate the detection results, we visualize misalignments using the state space graph (Fig.~\ref{fig:trajectory_visualization}). 
Each edge in the graph represents an action, and node sizes indicate the frequency of these actions in user trajectories. 

Edges highlighted in different colors correspond to detected misalignment types:
\begin{itemize}
    \item \textbf{Functional Inadequacy (Blue Edges)}: Sequences that lack an ideal action, requiring multiple inefficient steps.
    \item \textbf{User Unfamiliarity (Yellow Edges)}: Actions that could be performed more efficiently are not due to user errors.
    \item \textbf{Cognitive Dissonance (Red Edges)}: Incorrect trajectories that fail to reach the goal state.
\end{itemize}

\subsection{Misalignment Detection Algorithm}
\label{sec:misalignment_detection}

We introduce a lightweight yet structured misalignment detection algorithm based on popular states and ideal actions, which systematically detects and categorizes misalignments in human trajectories.
This algorithm evaluates each action sequence in a trajectory and classifies it as either an alignment transition or one of the three misalignment types defined earlier.
It is designed to operate directly on raw trajectory logs without requiring ground-truth intentions or manual annotations, making it both scalable and broadly applicable to various tasks and domains.

Alg.~\ref{alg:misalignment_detection} describes our heuristic approach, which systematically categorizes action sequences into alignment and misalignment types through three main steps.
(1) Identifying popular states as waypoints in trajectories,  
(2) Detecting action sequences between these waypoints, and  
(3) Classifying the sequences as aligned or misaligned based on deviations from ideal actions.

\begin{algorithm}
\caption{Misalignment Detection Algorithm}
\label{alg:misalignment_detection}

\textbf{Input:} Trajectories $\mathcal{T}_d$, State-Action Transition Function $f$, 

\phantom{\textbf{Input:}} Supported Action Set $\mathcal{A}_d$, Threshold Function $\theta$

\textbf{Output:} Identified Misalignments for Trajectories $\mathcal{M}_d$

$\mathcal{M}_d \gets \{\}$ \tcp{Initialize misalignment set}

\For{$t \in \mathcal{T}_d$}{

    \tcp{Step 1. Extracting Popular Nodes}
    $\mathcal{P}_d \gets \{\}$ 
    
    \For{$s_i \in t$}{
        \If{$N(s_i) \geq \theta(|\mathcal{T}_d|)$}{
            $\mathcal{P}_d \gets \mathcal{P}_d \cup \{(s_i, a_i)\}$
        }
    }

    \tcp{Step 2. Finding Misalignments}
    $M_t \gets \{\}$
    
    \For{$(s_i, a_i) \in \mathcal{P}_d$}{
        \If{$f(s_i, a_i) \neq s_{i+1}$}{
            \If{$\not\exists a \in \mathcal{A}_d$ such that $f(s_i, a) = s_{i+1}$}{
                $M_t \gets M_t \cup \{\textit{Functional Inadequacy in Tools}\}$
            }
            \Else{
                $M_t \gets M_t \cup \{\textit{User Unfamiliarity with Tools}\}$
            }
        }
    }

    \If{$s_n \neq \textbf{y}^{test}$}{
        $M_t \gets M_t \cup \{\textit{Cognitive Dissonance in Users}\}$
    }
    
    \tcp{Step 3. Aggregating Detected Misalignments}
    $\mathcal{M}_d \gets \mathcal{M}_d \cup \{M_t\}$
}

\Return $\mathcal{M}_d$

\end{algorithm}

\section{Misalignment Analysis}
\label{sec:analysis}
This section analyzes misalignments in human trajectories from ARCTraj across three hierarchical levels. 
We begin with an \textbf{action-level} analysis of the frequency and distribution of actions leading to specific states. 
Next, we examine \textbf{intention-level} misalignments between popular states. 
Finally, we explore \textbf{trajectory-level} relationships between misalignment types.

\subsection{Action-Level Analysis} 
This subsection analyzes action distributions across states using in-degree metrics to identify patterns commonly associated with different misalignment types.
In the state space graph (Fig.~\ref{fig:trajectory_visualization}), node size indicates the in-degree of a state, representing actions that lead to it from previous states.
Smaller average in-degrees suggest more diverse user strategies, fewer shared popular nodes, and thus a greater potential for misalignment across trajectories.

\paragraph{Average In-Degree} 
The in-degree distribution reveals meaningful differences in both task complexity and user behavior.
Tasks with smaller average in-degrees, summarized in Table~\ref{tab:bottom_10_node_size_tasks}, often involve low-level pixel adjustments or non-intuitive solution strategies.
For example, Task 17 required inefficient fill-in patterns across large grids, while Task 25 involved moving numerous pixels individually.
These tasks dispersed user actions across many intermediate states, leading to lower in-degree values and greater variability in the overall trajectories.

\begin{table}[htbp!]
\centering
\caption{Tasks with the bottom 10 average node sizes. These tasks involve complex patterns, multiple objects, or pattern filling, requiring diverse strategies.}
\begin{tabular}{@{}cccl@{}}
\toprule
\textbf{Rank} & 
\textbf{TaskID} & 
\textbf{Shortest Length} &  
\textbf{Description} \\
\midrule
 1 & 017c7c7b  & 4    & complex pattern \\
 2 & 0e206a2e  & 7    & multi-objects \\
 3 & 025d127b  & 1    & complex pattern \\
 4 & 0dfd9992  & 10+  & fill in the pattern \\
 5 & 05269061  & 3    & fill in the pattern \\
 6 & 1a07d186  & 10+  & complex pattern \\
 7 & 0962bcdd  & 2    & expand the pattern \\
 8 & 3428a4f5  & 4    & fill in the pattern \\
 9 & 09629e4f  & 5    & complex pattern \\
10 & 00d62c1b  & 1    & multi-objects \\
\bottomrule
\end{tabular}
\label{tab:bottom_10_node_size_tasks}
\end{table}

\paragraph{Key Insights from Action-Level Analysis} 
Low average in-degrees highlight Functional Inadequacies in Tools, where available actions fail to address task requirements efficiently. 
Pixel-level tasks, for instance, forced users to rely on repetitive actions, emphasizing the need for a more comprehensive action set.

With in-degree distributions, we identified key patterns highlighting Functional Inadequacies in Tools, particularly in tasks requiring repetitive or granular actions. 
These findings show how user behavior and tool limitations interact at a granular level. 
This action-level perspective sets the stage for intention-level analysis, where we investigate how sequences of actions between popular states contribute to broader misalignment patterns.

\subsection{Intention-Level Analysis}
This subsection examines misalignments at the intention level by analyzing action sequences between popular states, as identified using Alg.~\ref{alg:misalignment_detection}. 
Each intention corresponds to a sequence of actions connecting two consecutive popular states, enabling the categorization of misalignments within these segments.

\paragraph{Distribution of Misalignment Types}
Table~\ref{tab:intention_misalignment_distribution} presents the distributions of intentions and actions across different misalignment types. 
Aligned intentions dominate the dataset, comprising 91.11\% of all intentions. 
However, they represent only 49.57\% of the total actions, showing that misaligned intentions often require more extended or more redundant action sequences. 
Functional Inadequacy contributes most significantly to this imbalance, as it frequently involves highly repetitive pixel-level operations that substantially inflate the number of actions.

\begin{table}[htbp!]
\centering
\caption{Distribution of misalignment types at the intention level, showing their relative occurrence as a proportion of total intentions and actions. Misaligned intentions, although fewer, account for a significant share of suboptimal actions.}
\label{tab:intention_misalignment_distribution}
\begin{tabular}{@{}lcc@{}}
\toprule
\textbf{Misalignment Type} & \textbf{Intentions} & \textbf{Actions} \\ 
\midrule
User Unfamiliarity       & 2.31\%            & 11.30\%          \\  
Functional Inadequacy    & 4.15\%            & 27.54\%          \\  
Cognitive Dissonance     & 2.43\%            & 11.59\%          \\  
Not Misaligned           & 91.11\%           & 49.57\%          \\  
\bottomrule
\end{tabular}
\end{table}

\paragraph{Key Insights from Intention-Level Analysis}
\begin{itemize}
    \item \textbf{Aligned intentions dominate}: The high proportion of aligned intentions (91.11\%) demonstrates consistent user behavior across most tasks. 
    This suggests that human trajectories capture well-structured strategies, making them a reliable dataset for analyzing user behavior.

    \item \textbf{Functional Inadequacy drives inefficiency}: Despite representing only 4.15\% of intentions, Functional Inadequacy accounts for 27.54\% of all actions, revealing toolset limitations for tasks requiring repetitive or complex modifications. Addressing these inefficiencies through tool enhancements improves user performance.

    \item \textbf{Overlap in misalignment types}: User Unfamiliarity (2.31\%  and Cognitive Dissonance (2.43\%) exhibited similar intention proportions. 
    Also, their action proportions (11.30\% and 11.59\%) show similar patterns. 
    This overlap indicates potential shared causes but highlights the need for further investigation into how these misalignments differ in trajectory-level manifestations.
\end{itemize}

The intention-level analysis revealed that most user strategies align with ideal actions, though misaligned intentions lead to suboptimal behavior. This highlights how tool inefficiencies and user strategies manifest at the intention level. Building on this, the subsequent analysis examines the cumulative effects of misalignments across entire action sequences.

\subsection{Trajectory-Level Analysis}
\begin{figure}[htbp!]
\centering
\includegraphics[width=0.82\columnwidth]{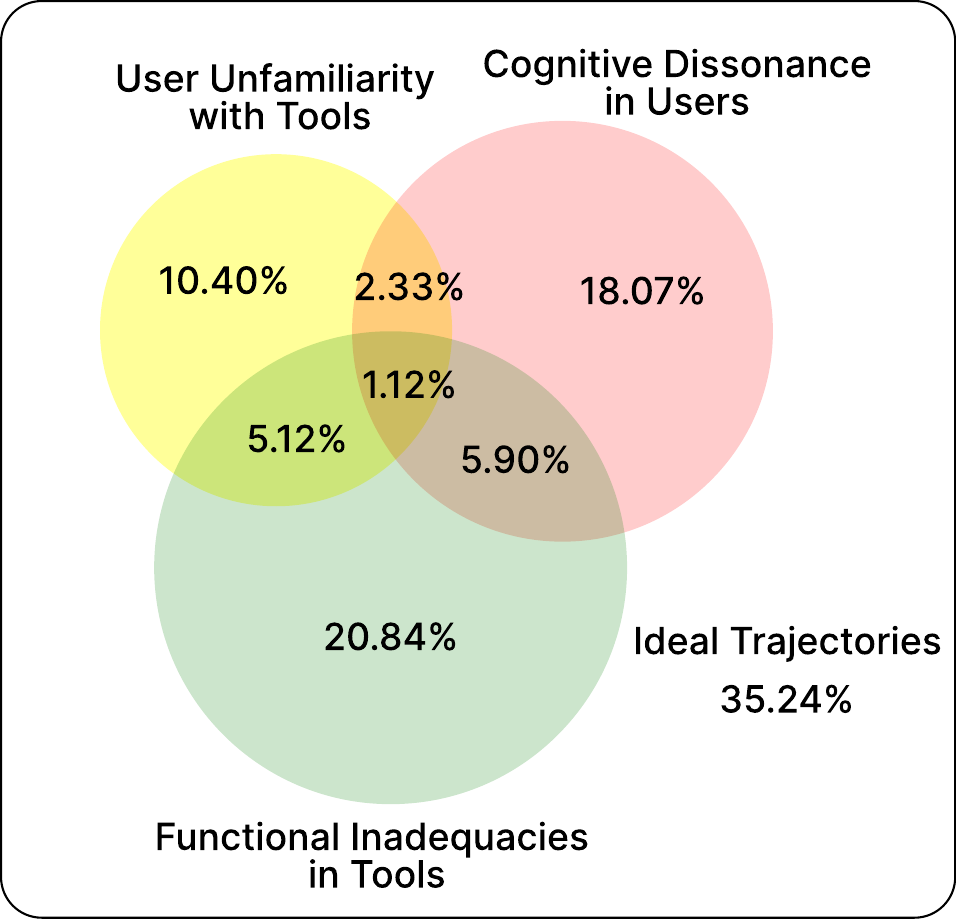}
\caption{A Venn diagram of misalignment types at the trajectory level, showing proportions of trajectories with different misalignment types and overlaps between them.}
\Description{A Venn diagram of misalignment types at the trajectory level, showing proportions of trajectories with different misalignment types and overlaps between them.}
\label{fig:trajectory_misalignment_venn}
\end{figure}

This subsection examines misalignments at the trajectory level by analyzing complete sequences of actions from the initial state to the final state. 
Fig.~\ref{fig:trajectory_misalignment_venn} illustrates the distribution of misalignments and their overlaps.

\paragraph{Key Insights from Trajectory-Level Analysis}
\begin{itemize}
    \item \textbf{Consistency in ideal trajectories}: The proportion of ideal trajectories (35.24\%) highlights structured user strategies, particularly for simpler tasks with intuitive solutions. This suggests that human trajectories capture meaningful behavior patterns essential for training AI systems to emulate effective approaches.

    \item \textbf{Prevalence of Functional Inadequacy}: Functional Inadequacy appeared in over 40\% of misaligned trajectories, underscoring limitations in the current action set. Tasks involving repetitive operations, such as pixel-level modifications, were particularly affected. Addressing these issues with tool enhancements could reduce misalignments and improve efficiency.

    \item \textbf{Minimal overlap between misalignment types}: Overlaps between misalignment types were minimal, especially between User Unfamiliarity and Cognitive Dissonance. This indicates distinct causes: User Unfamiliarity stems from inefficient tool usage, while Cognitive Dissonance reflects task misinterpretations or incomplete solutions. Differentiating these misalignments enables tailored interventions.
\end{itemize}

The trajectory-level analysis revealed that Functional Inadequacy is the most prevalent misalignment type, with minimal overlap between other misalignments. These findings suggest distinct causes for each misalignment type and reinforce the need to address tool limitations. The following section demonstrates how inferred intentions from trajectories enhance sequential modeling, providing a deeper understanding of user behavior and tool challenges.

\section{Evaluating Intention-Aligned Trajectories}
\label{sec:experiments}

\begin{figure*}[htbp!]
    \centering
    \includegraphics[width=0.24\textwidth]{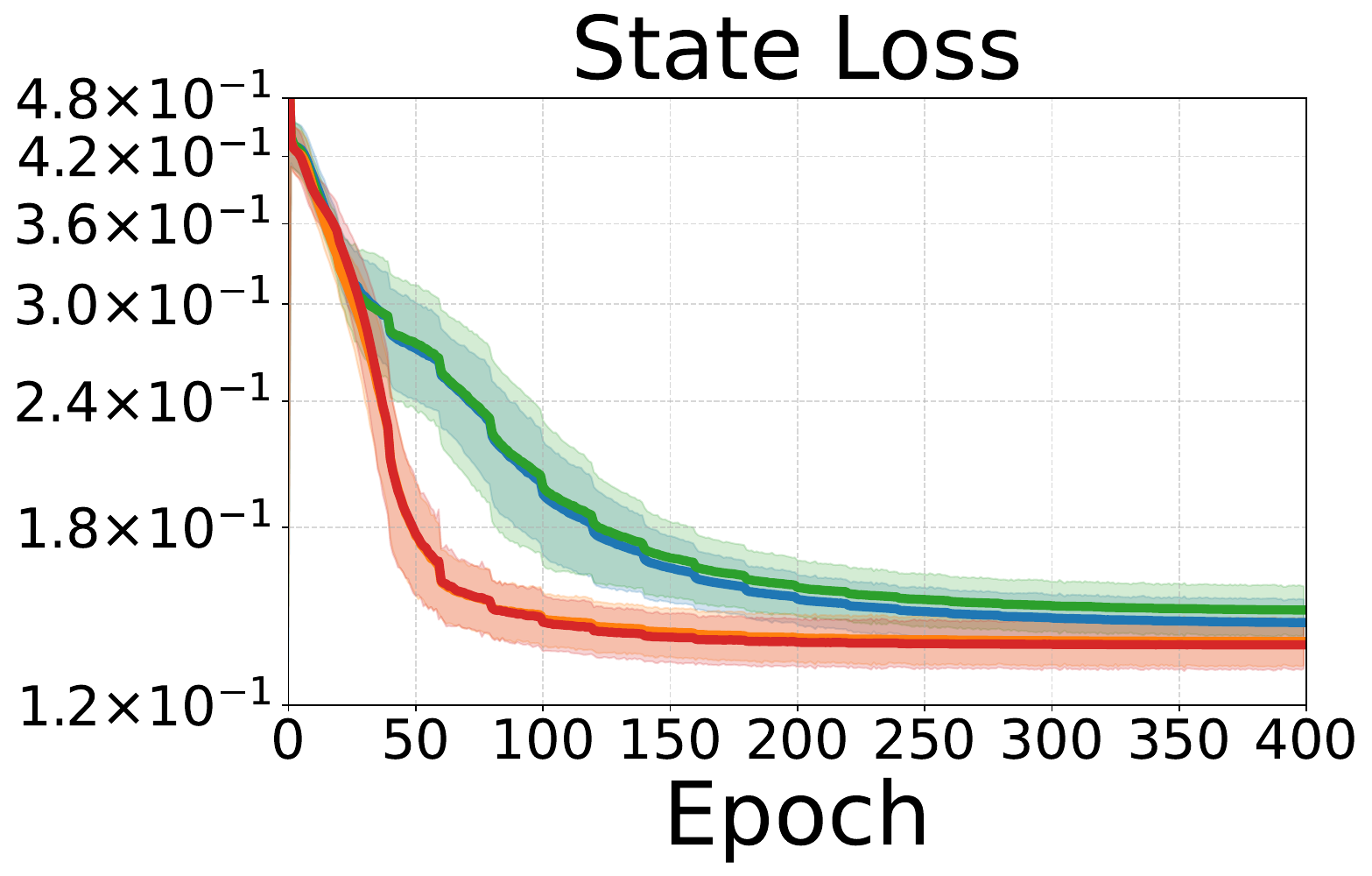}
    \includegraphics[width=0.24\textwidth]{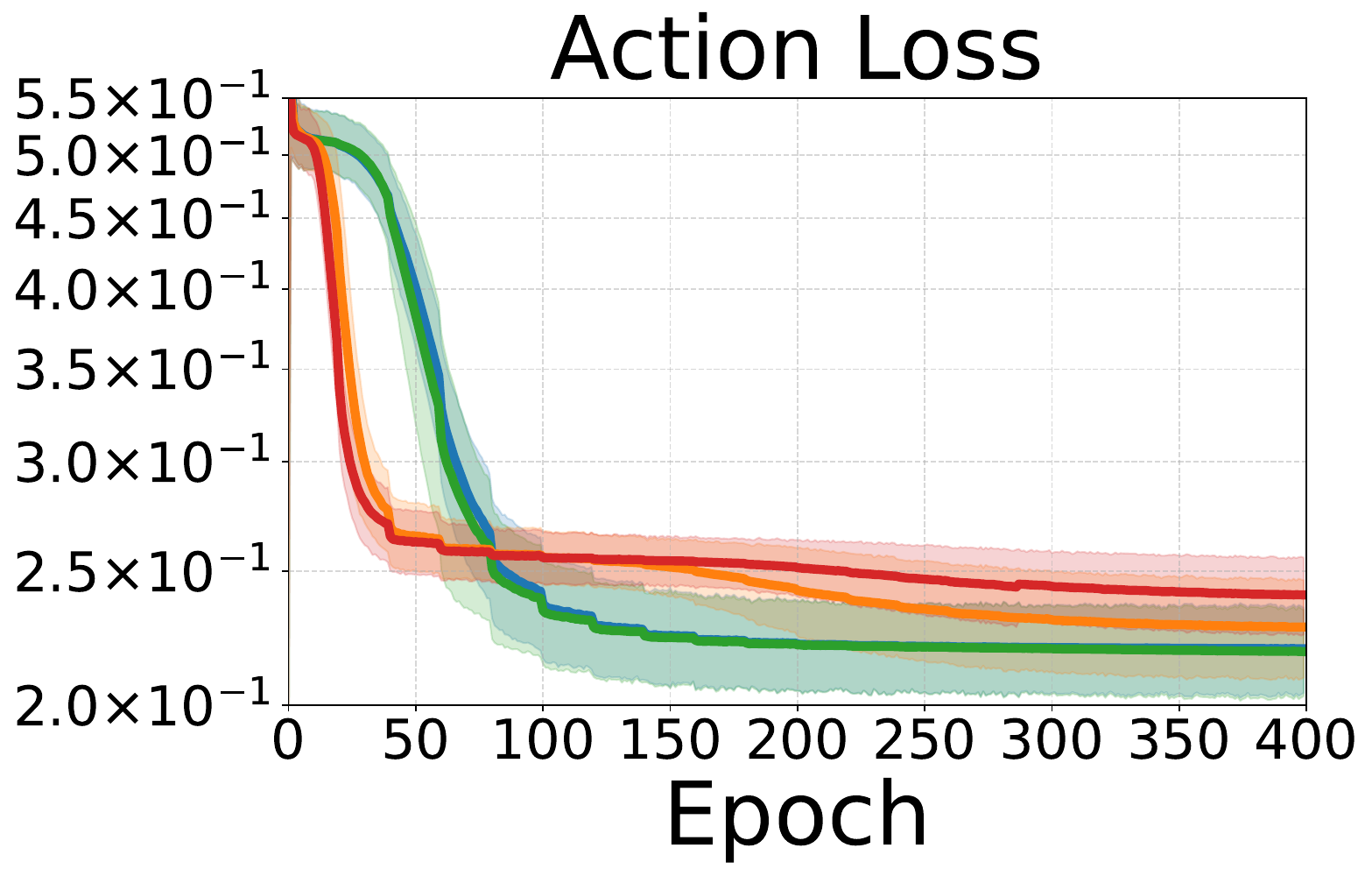}
    \includegraphics[width=0.24\textwidth]{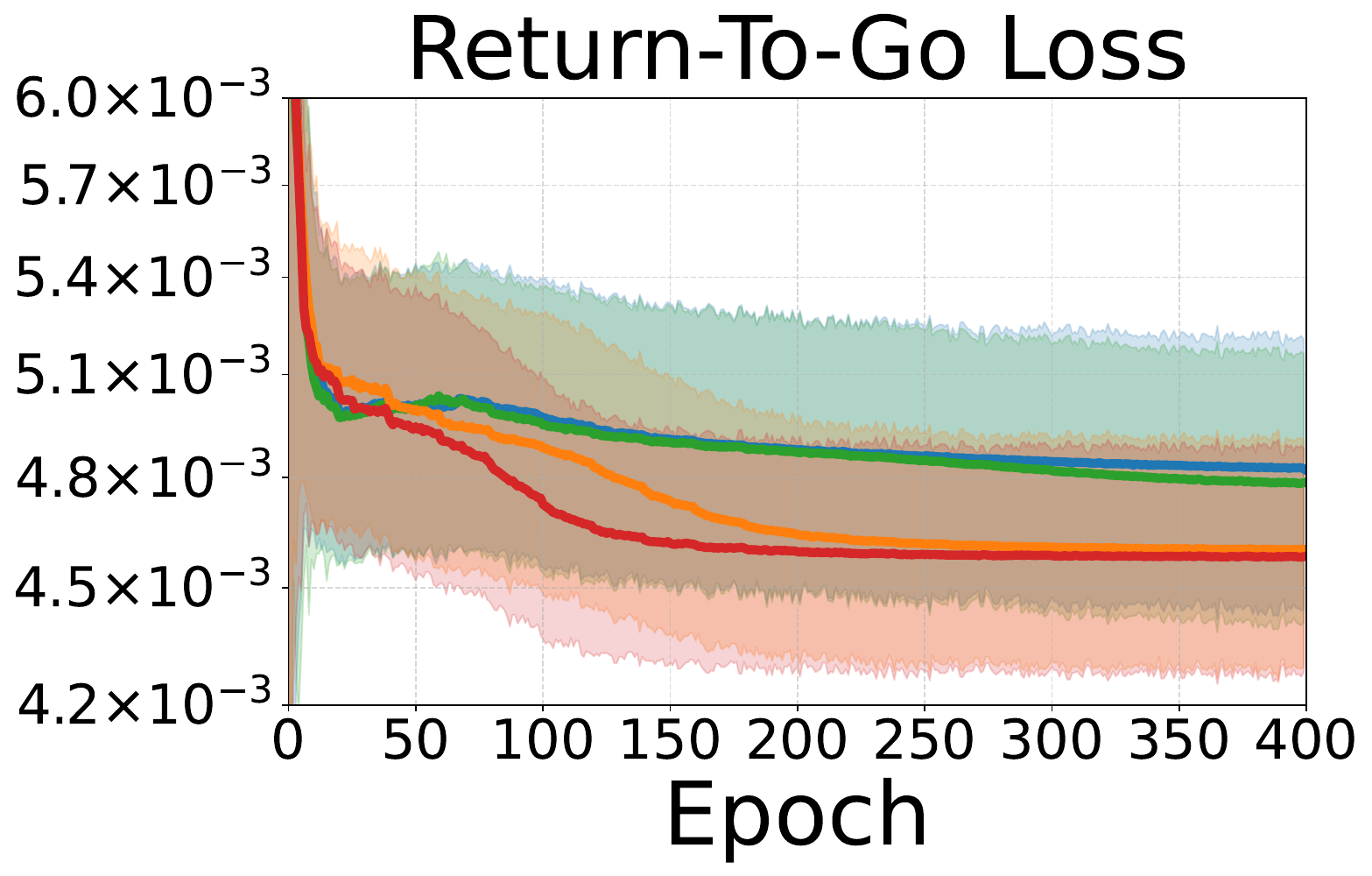}
    \includegraphics[width=0.24\textwidth]{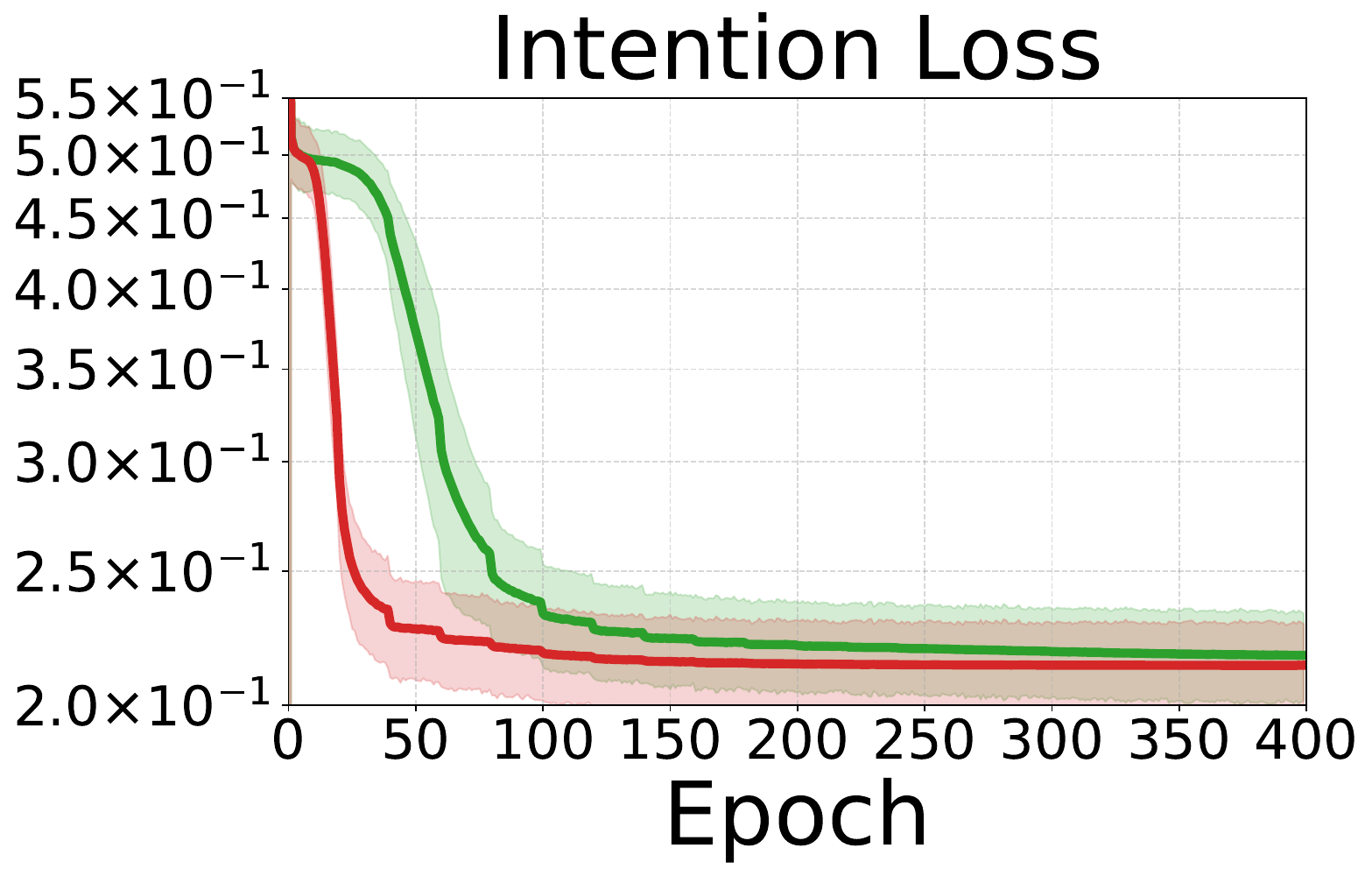}
    
    \includegraphics[width=0.7\textwidth]{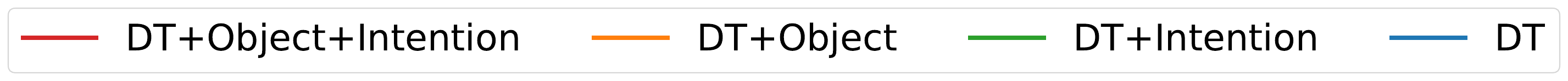}
    
    \caption{Training loss curves for different components. 
    (a) State loss measures the cross-entropy error for predicting the state grid. 
    (b) Action loss represents the cross-entropy error for action prediction. 
    (c) Return-to-go loss is calculated using MSE for reward estimation. 
    (d) Intention loss quantifies the cross-entropy error for inferred intentions.}
    \Description{Training loss curves for different components. State loss, Action loss, Return-to-go loss, and Intention loss.}
    \label{fig:loss_curves}
\end{figure*}

In the previous sections, we analyzed misalignments in human trajectories from ARCTraj and identified that a large portion, approximately 65\% according to Fig.~\ref{fig:trajectory_misalignment_venn}, exhibits some form of suboptimal behavior. This observation underscores the potential benefits of explicitly addressing these misalignments to enhance the efficiency and generalization of trajectory-based learning systems.

Motivated by this, we propose a method to infer implicit user intentions by identifying transitions between popular states that likely represent meaningful steps in human problem-solving. We incorporate these inferred intentions as auxiliary supervision in model training and evaluate whether doing so leads to improved task-solving performance and a better alignment with structured reasoning patterns in human demonstrations.

\subsection{Intention Prediction Algorithm}
We present Alg.~\ref{alg:intention_prediction}, which formalizes the process of intention prediction by leveraging popular states as key decision points. Our key assumption is that most human demonstrations follow correct task-solving procedures and that popular states represent crucial intermediate waypoints in these processes.

\begin{algorithm}
\caption{Intention Prediction Algorithm}
\label{alg:intention_prediction}

\textbf{Input:} Trajectories for Solving task $\mathcal{T}_d = (s_0, a_0, \cdots, s_n)$,

\phantom{\textbf{Input:}} Threshold Function $\theta$

\textbf{Output:} Task Trajectories with Assigned Intentions $\mathcal{T}$

$\mathcal{P} \gets \{\}$ \tcp*[r]{Initialize popular states set}

\tcp{Step 1. Identify Popular States}
\For{$\tau_d \in \mathcal{T}_d$}{
    \For{$s_i \in \tau_d$}{
        \If{$N(s_i) \geq \theta(|\mathcal{T}|)$}{
            $\mathcal{P} \gets \mathcal{P} \cup \{s_i\}$ ;
        }
    }
}

\tcp{Step 2. Assign Intention Edges}
\For{$\tau_d \in \mathcal{T}$}{
    $s \gets \tau_d[s_0]$ ;
    \For{$s_i, a_i \in \tau_d$}{
        \If{$s_i \in \mathcal{P}$}{
            \For{$s_j, a_j \in \tau_d[s: s_i]$}{
                $a_j[intention] \gets (s, s_i)$ ;
            }
            $s \gets s_i$ \tcp*[r]{Update current popular state}
        }
    }
}

\Return $\mathcal{T}_d$

\end{algorithm}

Building on this assumption, we hypothesize that user intentions emerge through transitions between popular states. For each trajectory $\tau_d$, we annotate every action sequence $a_{i:j}$ occurring between two such states $(s_i, s_j)$ with an intention tuple $(s_i, s_j)$. This pair defines a conceptual ideal edge $a^\star_{i:j}$ that abstracts the user’s intended transformation.

To assess the value of this information, we augment training data with these inferred intentions. Inspired by prior work that utilizes human trajectories for training~\cite{park2023unraveling}, we evaluate whether this additional supervision improves model performance on task-solving benchmarks.

\subsection{Experiment}
To evaluate the impact of intention supervision, we incorporated intention annotations into an existing Decision Transformer (DT) framework~\cite{park2023unraveling}. Using a predefined transformation policy, the original DT was trained on expert trajectories derived from augmented input-output examples. The model learns to predict state, action, return-to-go, and timestep by attending to the sequence of trajectories. Prior work has shown that augmenting input states with object-level features improves performance in structured tasks.

We extended this framework by integrating intention information. Specifically, the model was modified to handle intention-labeled transitions and was trained to predict intention tuples connecting popular states. We implemented this by adding an auxiliary classification head and corresponding loss term dedicated to intention prediction. We hypothesize that this supervision encourages the model to learn structured, goal-oriented patterns of reasoning rather than simply mimicking surface-level actions.

\paragraph{Baselines}
In this study, we experimented with four variations of the \textbf{Decision Transformer (DT)} to solve the ARC task. Each model takes state, action, and return-to-go as inputs and predicts future values accordingly, learning from temporally ordered human demonstration sequences. We extended the input representations by incorporating object-level features and intention annotations to assess better how structured information impacts learning. These enhancements were designed to guide the model toward more interpretable, goal-aligned reasoning paths and evaluate whether they improve generalization across diverse task types.

\begin{itemize}
\item{\textbf{DT}:} The standard DT model was modified to predict return-to-go, state, and action. Since the state in human trajectories is represented as a 2D grid of pixel values ranging from 0 to 9, predicting the state is equivalent to predicting every pixel in the grid. The input to the DT model consists of (return-to-go, state, action) sequences from time step \(t\) to \(t+T-1\), and the model is trained to predict the corresponding values from \(t+1\) to \(t+T\).

\item{\textbf{DT + Object}:} In this variant, we introduced object information by applying a heuristic algorithm to infer objects based on pixel grouping in each state. Instead of storing the object information as a complete state representation, we embedded it to match the dimensions of return-to-go and action, ensuring efficient use of model capacity.

\item{\textbf{DT + Intention}}: We proposed using Alg.~\ref{alg:intention_prediction} to introduce intention information during training. However, since this information is unavailable during testing, the model was trained to infer it implicitly. The intention was encoded as an integer, representing the popular states that define an ideal action category.

\item \textbf{{DT + Object + Intention}:} This model incorporated both object and intention information. It achieved the highest performance among all variations, demonstrating the effectiveness of leveraging structured representations in the ARCTraj dataset.
\end{itemize}

\paragraph{Losses}
For training, we used \textbf{cross-entropy loss} for state, action, and intention predictions, as their values are categorical. In contrast, \textbf{mean squared error (MSE) loss} was applied to return-to-go, which is a continuous value in the range of 0 to 1. The \textbf{total loss} was computed as the sum of all individual components, but only the component-wise losses are visualized in Fig.~\ref{fig:loss_curves}. Notably, models without intention information had a lower total loss since they did not include an intention loss term.

\subsection{Result}
\begin{figure}[htbp!]
    \centering
    \includegraphics[width=0.85\columnwidth]{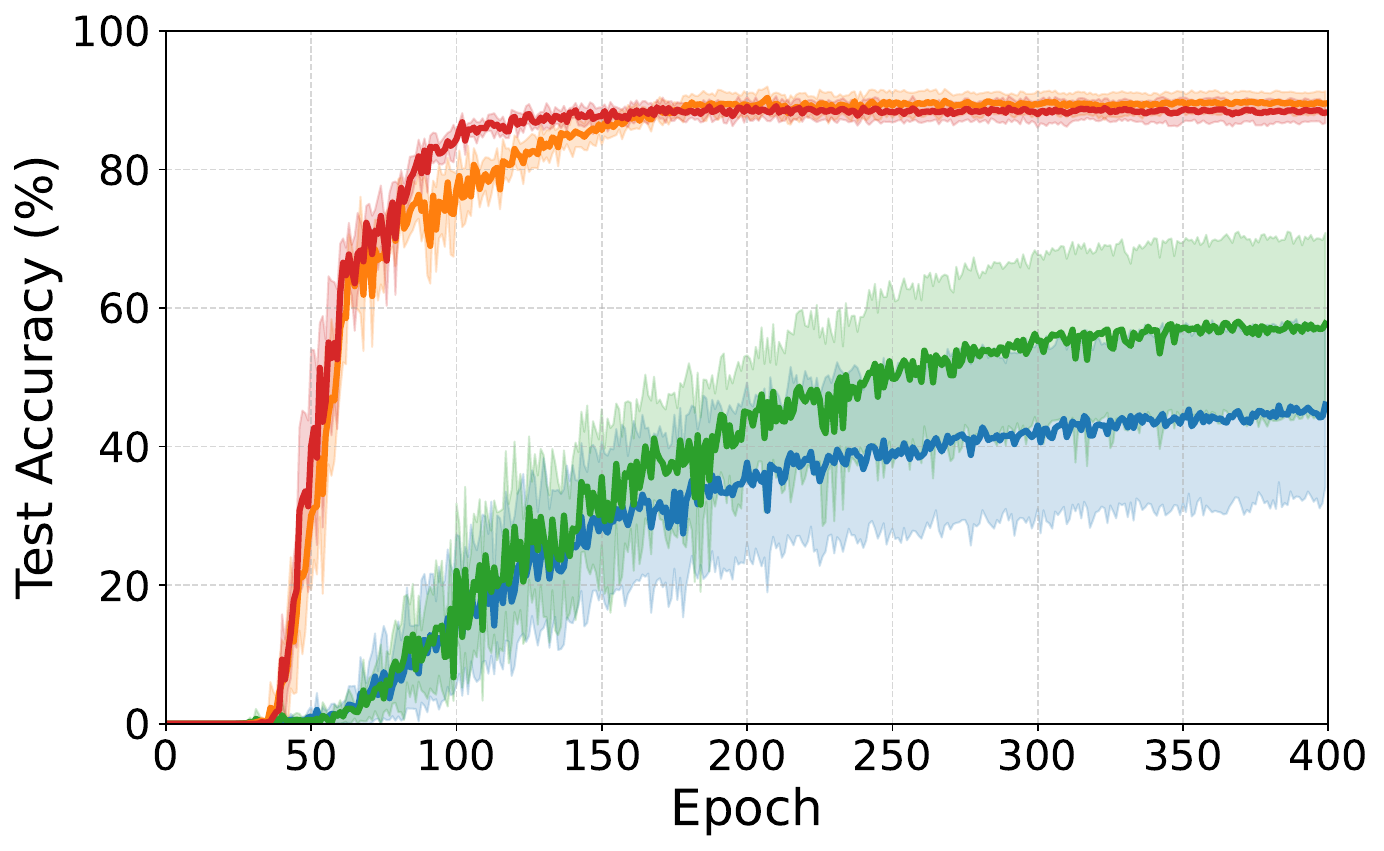}
    \includegraphics[width=0.85\columnwidth]{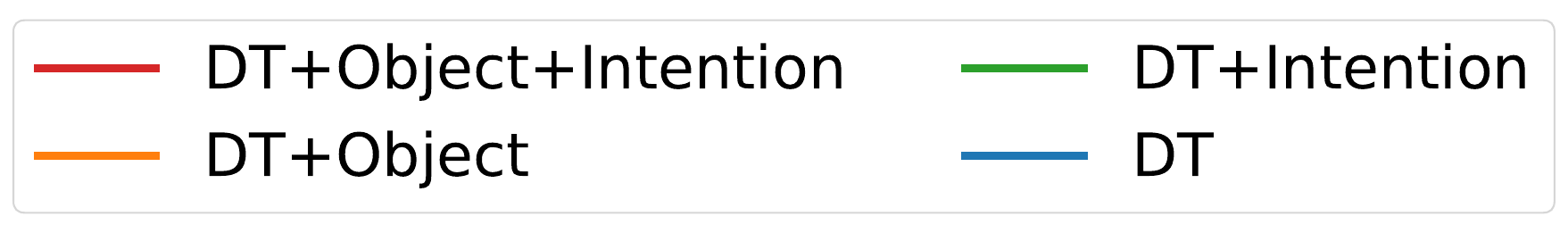}
    
    \caption{Test accuracy comparison of different models. The DT + Object + Intention model outperforms others, demonstrating the effectiveness of intention alignment. However, adding intention supervision to the baseline DT model without object features leads to a performance drop, highlighting the importance of combining object and intention information for optimal learning.}
    \Description{Test accuracy comparison of different models. The DT + Object + Intention model outperforms others, demonstrating the effectiveness of intention alignment. However, adding intention supervision to the baseline DT model without object features leads to a performance drop, highlighting the importance of combining object and intention information for optimal learning.
    }
    \label{fig:test_accuracy}
\end{figure}

Our results are summarized in Fig.~\ref{fig:test_accuracy}, clearly demonstrating that incorporating intention supervision significantly enhances task-solving performance. Specifically, adding intention information to the existing model with object features (DT + Object) resulted in an average performance improvement of \textbf{5.85\%} across four evaluation datasets, each consisting of 2,000 augmented ARC tasks. Despite the already strong baseline performance of DT + Object (83.59\%), integrating intention alignment further boosted accuracy to nearly 90\%, providing strong empirical evidence that intention-based supervision meaningfully improves model learning outcomes.

However, as seen in the training loss curves, training with intentional supervision required more samples and training time, suggesting increased learning complexity, which is expected since the model must jointly learn both object-level and intention-based reasoning in a coordinated fashion. Notably, when intention information was added to the baseline model without object features (DT + Intention), the model exhibited reduced efficiency in leveraging intention cues, leading to a measurable drop in performance rather than any meaningful improvement. This indicates that intention alignment alone is insufficient without a structured and interpretable internal representation of object-level information.

These findings highlight the importance of integrating structured object representations with intention-based supervision to enhance trajectory learning. While intention alignment enhances reasoning over key transitions, its effectiveness depends on the availability of complementary features such as object information. Future research could explore more efficient training strategies or explicit intention annotation methods to refine this approach, such as curriculum learning over intentions or semi-supervised annotation schemes for real-world applications.

\subsection{Ablation}
\label{sec:ablation}

To evaluate the robustness and effectiveness of inferred intention signals, we conducted ablation experiments using dummy variants.
Specifically, we compared the predictive accuracy of models trained under four settings:
(1) no intention information,
(2) fixed dummy values (zeros),
(3) randomly sampled intentions,
and (4) predicted intentions from Alg.~\ref{alg:intention_prediction}.

As shown in Table~\ref{tab:ablation}, only the model trained with predicted intentions significantly outperforms the baseline.
The DT + Object + Intention (predicted) model achieves 89.44\% accuracy, compared to 83.59\% for DT + Object without intentions.
This demonstrates that learning with inferred, aligned intentions contributes to performance improvements.

By contrast, feeding the model fixed dummy values (e.g., zero vectors) yields only marginal gains in performance.
This suggests that the model effectively learns to ignore uninformative or redundant intention signals during training.
More notably, when intention labels are randomly sampled from unrelated tasks, performance drops significantly to 1.10\%.
This highlights that incorrect or misleading intention inputs can actively harm learning by introducing contradictory supervision that confuses the model’s internal representations.

\newpage
\begin{table}[htbp!]
    \centering
    \caption{Ablation study on DT + Object with different intention sources. Only predicted intentions improve accuracy.}
    \label{tab:ablation}
    \begin{tabular}{lcc}
        \toprule
        \textbf{Model Variant} & \textbf{Intention Type} & \textbf{Accuracy} \\
        \midrule
        DT + Object & None & 83.59\% \\
        + Intention (0) & Fixed dummy & 84.12\% \\
        + Intention (random) & Random dummy & 1.10\% \\
        + Intention (predicted) & Alg.~\ref{alg:intention_prediction} & \textbf{89.44\%} \\
        \bottomrule
    \end{tabular}
\end{table}

These findings support two key conclusions.
First, intention supervision can improve task-solving performance, but only if the signals are meaningful and aligned with human behavior.
Second, injecting low-quality or noisy intentions is not only ineffective but actively harmful.
Thus, the benefits of intention alignment depend heavily on the quality and structure of the inferred annotations.

\section{Conclusion}

This study introduced a structured approach for understanding and improving human-like reasoning by analyzing user trajectories from the ARCTraj dataset. By systematically identifying and formalizing three types of trajectory misalignments (i.e., Functional Inadequacies in Tools, User Unfamiliarity with Tools, and Cognitive Dissonance in Users), we provided a comprehensive and interpretable framework for diagnosing and explaining errors in human problem-solving behavior.

We proposed an intention prediction algorithm based on transitions between popular states to mitigate these misalignments. This method enabled us to annotate trajectories with high-level intention structures that reflect user reasoning and approximate their latent goals. Experimental results demonstrated that incorporating these inferred intentions into model training significantly enhances task-solving performance and promotes structured learning. Notably, models that integrated both object-level features and intention supervision (DT + Object + Intention) achieved nearly 90\% accuracy, outperforming all other baselines and demonstrating the benefit of combining multiple structural signals.

These findings suggest that intention-aligned supervision helps models acquire structured reasoning patterns beyond surface-level execution. However, we also observed that intention signals alone, without complementary object representations, can degrade performance. This emphasizes the importance of combining multiple levels of structure when modeling human reasoning.

Future work will explore more expressive models for intention inference, such as graph neural networks or transformers, to better capture latent structure in demonstrations. Improving interface design and refining evaluation via state space analysis may also offer more effective diagnostics. As datasets like ARCTraj grow, aligning intention-based representations with learning goals will be key to building models that generalize systematic reasoning across diverse domains.

\begin{acks}
This work was supported by IITP (RS-2023-00216011, RS-2024-004450807, No. 2019-0-01842), NRF (RS-2024-00451162, RS-2024-00454000), and GIST (Postdoc Value-up) grants funded by the Ministry of Science and ICT, Korea. 
\end{acks}

\newpage

\bibliographystyle{ACM-Reference-Format}
\bibliography{99_reference}

\appendix
\iffalse
\section{Appendix Overview}

This appendix provides additional details to support the methodology and analysis presented in the main paper, divided into three sections. First, \textbf{Characteristics of Human Trajectories} describes the unique properties of ARCTraj trajectories~\cite{kim2025arctraj}, including their brevity, temporal dependency, and multi-user variability. Second, \textbf{State Representation of Human Trajectories} outlines the complete structure of the ARCTraj dataset's state representation. Third, a \textbf{Case Study on Task 53b68214} illustrates how the proposed visualization tool and detection algorithm reveal misalignment patterns and user strategies. Supplementary figures and logs are included to support reproducibility and qualitative interpretation.
\fi

\section{Characteristics of Human Trajectories}

The ARCTraj dataset captures user interactions during ARC task solving, offering distinct properties that influence learning strategies. Trajectories are relatively short, typically 10–30 actions, which limits temporal context compared to longer reinforcement learning episodes. This brevity simplifies inspection and reduces computational cost, although it makes intention inference more difficult. Each action depends on previous ones, reflecting the sequential and deliberative nature of human reasoning. Misalignments often stem from inconsistencies across steps, so models must be able to capture temporal structure to accurately interpret behavior.

Additionally, each task includes trajectories from approximately 25 users, introducing a range of strategic variations. While this enriches the dataset, it also challenges models to generalize across different approaches without overfitting. Actions are compositional, consisting of an \texttt{Operation} (e.g., \texttt{rotate}, \texttt{paint}) and a \texttt{Selection} over the grid. This structure increases expressivity but also adds modeling complexity, as suboptimal operations or incorrect selections may indicate user errors or misunderstandings. These combined characteristics, including brevity, sequentiality, user diversity, and compositionality, serve as the foundation for our analysis and modeling assumptions.

\section{State Representation of Human Trajectories}

Each state in the human trajectories from the ARCTraj dataset comprises several components that describe the task environment and user interactions. These include the visible grid, object information, background layout, and clipboard contents. In this study, however, we focus solely on the \textbf{grid} representation, which captures the essential information required for analyzing task-solving behavior and misalignments.

Although our modeling relies solely on the grid state for simplicity and generality, the complete state representation comprises several distinct components that provide a richer, more nuanced view of user behavior. These include:

\begin{itemize}
\item \textbf{Grid}: A 2D pixel array representing visible task state, including object layout and background colors.
\item \textbf{Objects}: Defined by user selections, with attributes such as shape and position, which help capture structured changes.
\item \textbf{Background}: Unselected grid regions that offer structural context for interpreting task layout.
\item \textbf{Clipboard}: Stores temporarily selected regions for operations such as \texttt{copy} and \texttt{paste}.
\end{itemize}

\section{Case Study: Task 53b68214}

\begin{figure}[htbp!]
\centering
\includegraphics[width=\columnwidth]{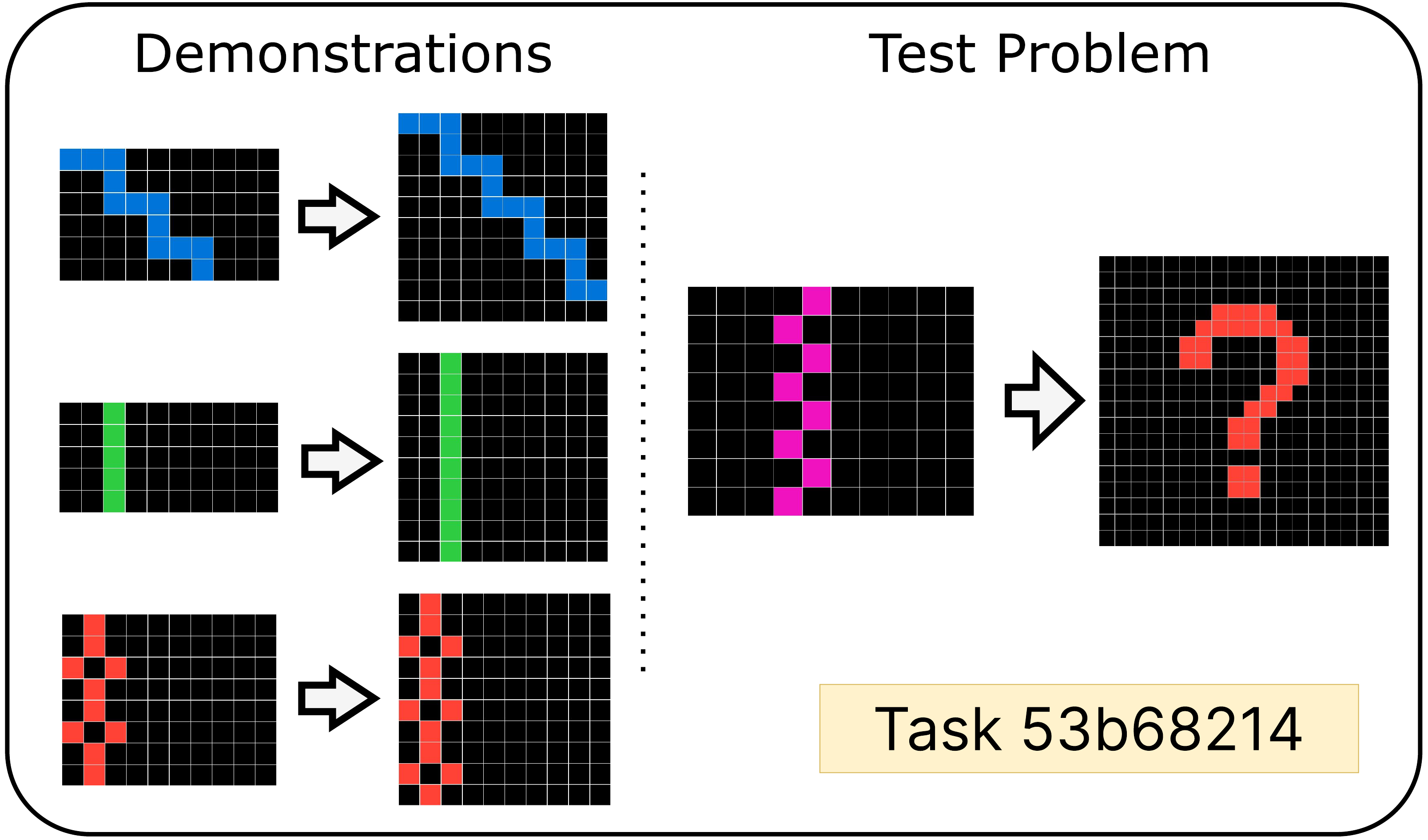}
\caption{ARC Task 53b68214. The rule is to resize the grid vertically to 10 and then fill the expanded region with the original grid pattern.}
\Description{ARC Task 53b68214. The rule is to resize the grid vertically to 10 and then fill the expanded region with the original grid pattern.}
\label{fig:case_study_arc_task}
\end{figure}

\subsection{Task Overview and Misalignment Examples}

To demonstrate how misalignments can be identified using our visualization tool, we present a case study of ARC Task 53b68214, shown in Fig.~\ref{fig:case_study_arc_task}. This task requires users to expand the vertical size of the grid to 10 rows and then add a magenta zigzag pattern to the newly appended lower region. Despite the rule's apparent simplicity, users must correctly understand and execute two distinct subgoals: resizing the grid to the appropriate dimensions and generating a non-trivial, spatially aligned color pattern. The final correct output is a $10 \times 10$ grid, where the lower portion contains the added magenta structure, aligned with task constraints. While the optimal solution requires only a few precise and well-ordered operations, users often diverged from this intended path, resulting in observable misalignment patterns across their trajectories.

By analyzing the trajectory graph from ARC Task 53b68214 (Fig.~\ref{fig:case_study_visualization}), we identified concrete instances for each of the three misalignment categories introduced earlier. These cases show how diverse user behaviors manifest as structural patterns in the graph, highlighting the value of our misalignment detection framework.

\begin{itemize}
    \item \textbf{Functional Inadequacies in Tools}:  
    Although the task can technically be solved using the available set of operations, some users struggled to efficiently replicate the color pattern after resizing. This is partly due to the lack of a grid-preserving \texttt{resize} operation that automatically fills the expanded space based on prior patterns. While adding such an operation could improve efficiency, it may only apply to a narrow set of tasks and could reduce generalizability.

    \item \textbf{User Unfamiliarity with Tools}:  
    Several users exhibited inefficient action sequences involving the repetitive use of the \texttt{paint} tool to manually color individual pixels, despite the system’s ability to support multi-pixel painting through selection. This suggests a lack of awareness about more efficient interaction techniques. These inefficient trajectories are primarily clustered on the left portion of the graph, where multiple red nodes indicate failed attempts, including frequent back-and-forth edits and excessive operations.

    \item \textbf{Cognitive Dissonance in Users}:
    Many users overlooked resizing the grid to a $10 \times 10$ configuration, either skipping the step entirely or resizing incorrectly. In addition, some trajectories exhibit loops or cycles, such as performing an action and then immediately undoing it, which reflects user hesitation, low confidence, or task confusion. These behaviors are common in the top and bottom regions of the graph and often precede final incorrect submissions, highlighting difficulties in understanding task requirements.
\end{itemize}

\begin{figure}[htbp!]
\centering
\includegraphics[width=\columnwidth]{Figures/graph_task124.pdf}
\caption{Visualization of user trajectories for Task 53b68214. Each node shows a distinct grid state with its visual overlay. Edges indicate user actions, and node colors mark initial (blue), correct (green), incorrect (red), and intermediate (gray) states. Clusters and edge density highlight common paths, repeated strategies, and frequent misalignments across multiple users.}
\Description{Visualization of user trajectories for Task 53b68214. Each node shows a distinct grid state with its visual overlay. Edges indicate user actions, and node colors mark initial (blue), correct (green), incorrect (red), and intermediate (gray) states. Clusters and edge density highlight common paths, repeated strategies, and frequent misalignments across multiple users. This graph offers a visual summary of how humans approach the task and where their reasoning diverges.}
\label{fig:case_study_visualization}
\end{figure}

\subsection{Visualization Details}

We developed an interactive HTML-based graph visualization tool to support large-scale, structured analysis of human trajectories across ARC tasks. For each task, the tool automatically generates a standalone HTML file named after its ARC Task ID (e.g., \texttt{53b68214.html}), which visually presents all collected user trajectories as a unified, interactive directed graph. In this graph, each node corresponds to a unique grid configuration encountered at some point in a trajectory, and each edge represents an \texttt{operation} that transitions between states. Users can click on nodes to inspect the grid as a 2D emoji array and hover over edges to view the associated operation type and context.

\begin{figure}[htbp!]
\centering
\includegraphics[width=\columnwidth]{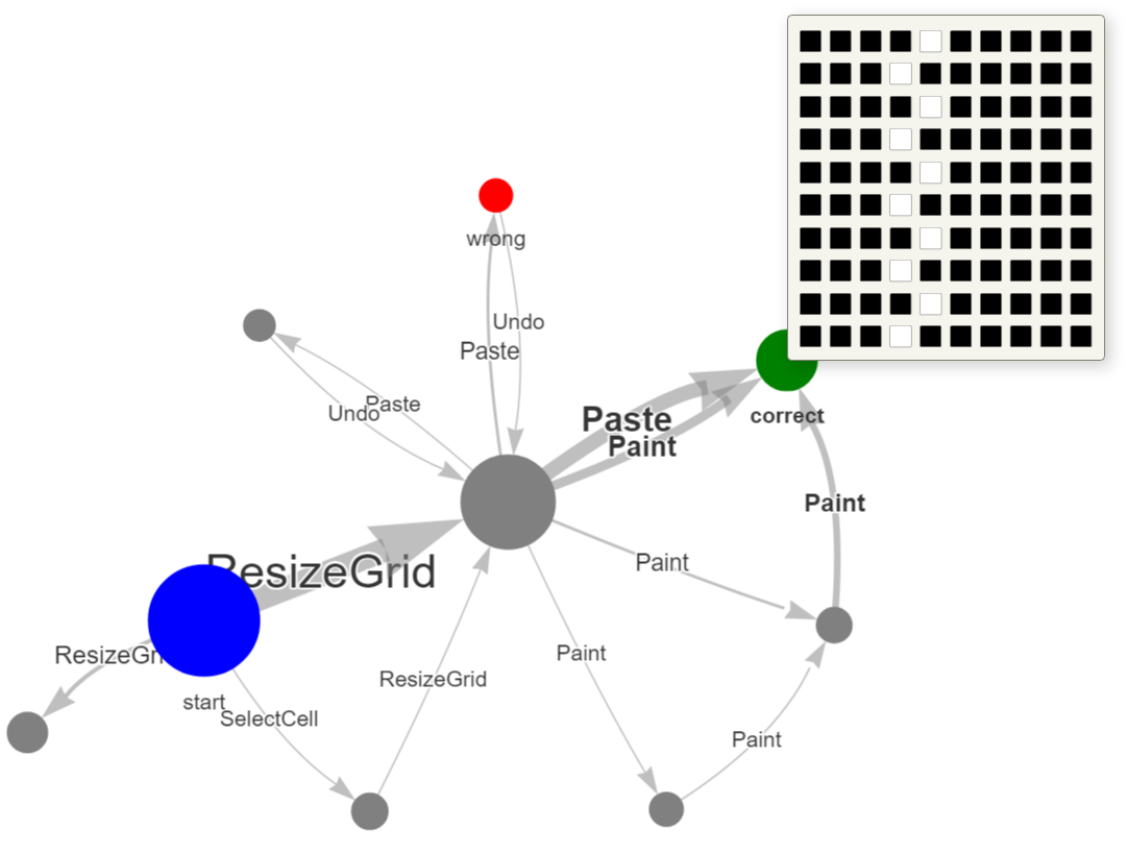}
\caption{Visualization of user trajectories for Task 53b68214 (simplified graph). Nodes represent grid states without selection information, enabling a clearer view of state transitions. Clicking a node reveals its grid as a 2D emoji array. Colors indicate initial (blue), correct (green), incorrect (red), and intermediate (gray) states, supporting intuitive inspection of user strategies.}
\Description{Visualization of user trajectories for Task 53b68214 (simplified graph). Nodes represent grid states without selection information, enabling a clearer view of state transitions. Clicking a node reveals its grid as a 2D emoji array. Colors indicate initial (blue), correct (green), incorrect (red), and intermediate (gray) states, supporting intuitive inspection of user strategies.}
\label{fig:case_study_visualization_without_selection}
\end{figure}

To improve clarity, our main visualization omits the \texttt{selection} component and focuses on operation types such as \texttt{paint}, \texttt{rotate}, and \texttt{paste}. This simplification reduces visual clutter by minimizing redundant nodes and edges that arise from selection-level variations. Node colors indicate semantic roles: blue for the initial state, green for the correct answer, red for incorrect submissions, and gray for intermediate states. This concise view enables intuitive navigation and comparison of strategies, revealing common behaviors and divergence points across user trajectories (see Fig.~\ref{fig:case_study_visualization_without_selection}).

\begin{figure}[htbp!]
\centering
\includegraphics[width=\columnwidth]{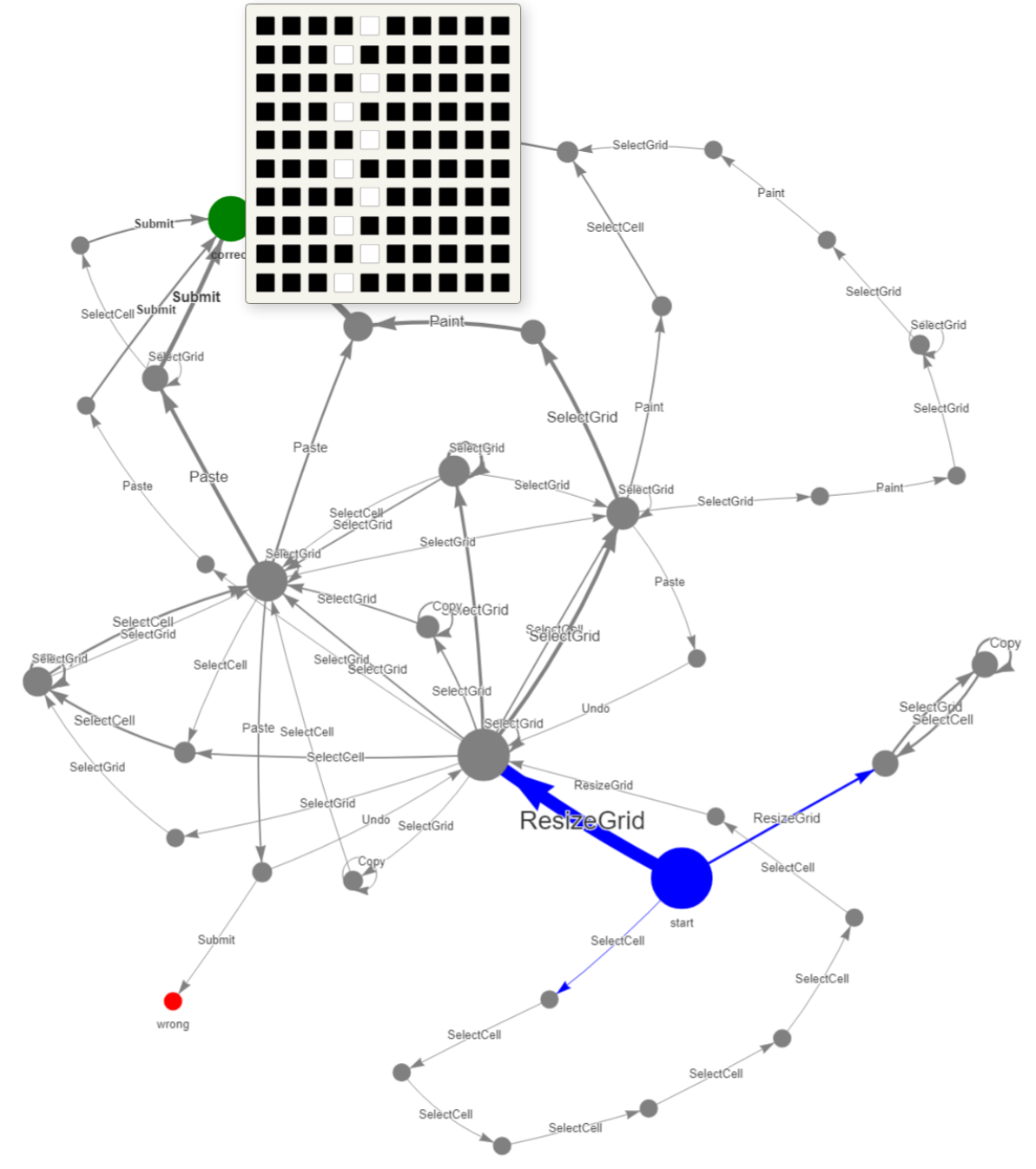}
\caption{Full-detail visualization of Task 53b68214 including selection. Increased node and edge complexity limited its utility for structural analysis.}
\Description{Full-detail visualization of Task 53b68214 including selection. Increased node and edge complexity limited its utility for structural analysis.}
\label{fig:case_study_visualization_with_selection}
\end{figure}

In contrast, we also implemented a full-detail version with both \texttt{operation} and \texttt{selection} in nodes and edges (Fig.~\ref{fig:case_study_visualization_with_selection}). While this variant enables fine-grained tracking, it greatly increases node count and adds many self-loops, complicating the state space graph. This complexity hindered pattern extraction and misalignment analysis. We therefore focused on the simplified version without selection, which offered a more tractable and interpretable view of user trajectories.

\subsection{Misalignment Detection}

\begin{figure}[htbp!]
\centering
\includegraphics[width=\columnwidth]{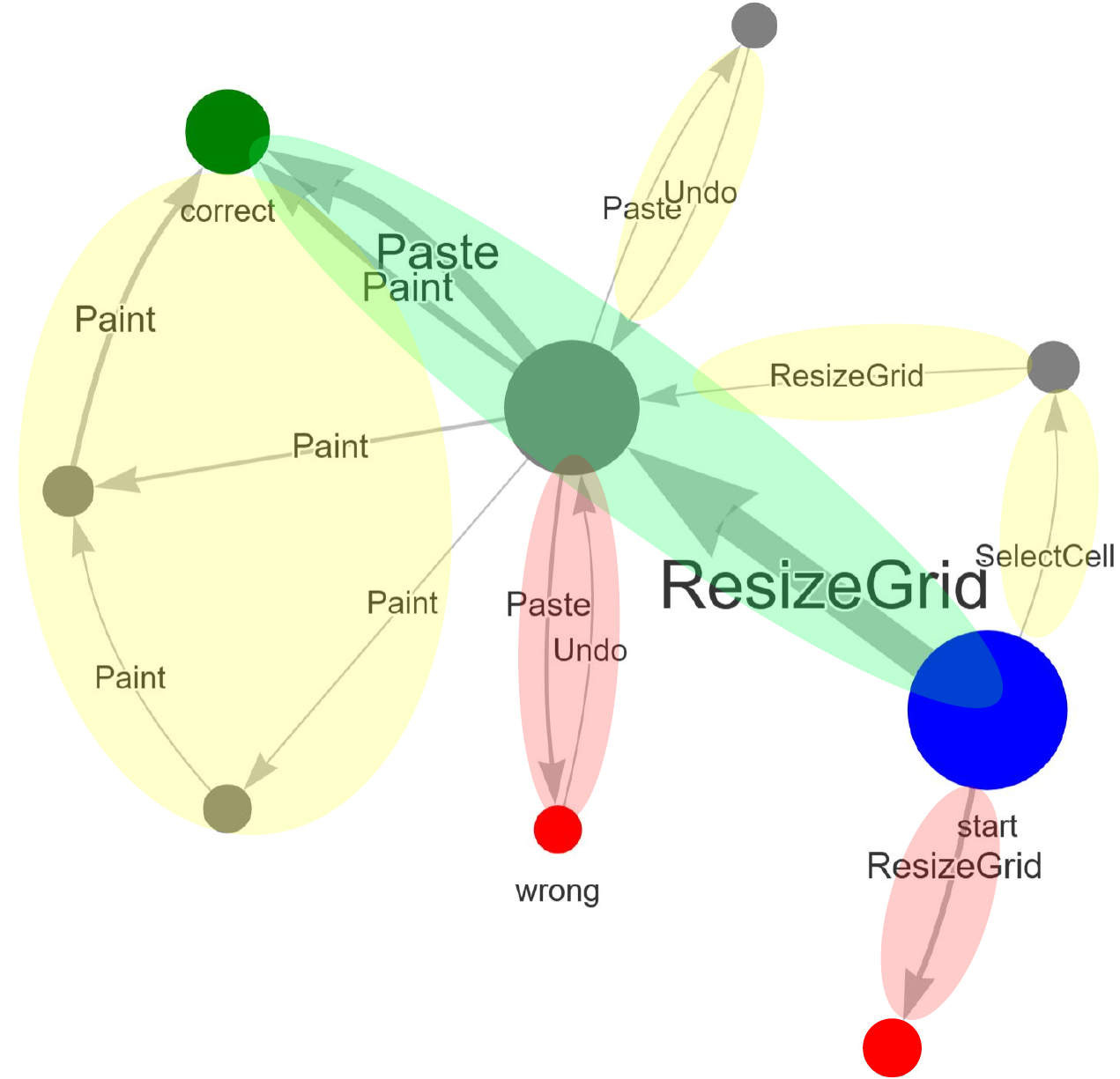}
\caption{Sub-trajectories in ARC Task 53b68214, where the proposed Misalignment Detection Algorithm (Alg.~\ref{alg:misalignment_detection}) identifies types of user misalignment. Red-circled regions indicate \textbf{Cognitive Dissonance in Users}, typically involving cyclic behavior such as undoing actions or navigating without a clear goal. Yellow circles highlight \textbf{User Unfamiliarity with Tools}, where users applied redundant sequences (e.g., repeated paint operations) to achieve outcomes that simpler actions could accomplish. Green ovals mark \textbf{Functional Inadequacies in Tools}, where limitations (e.g., no resize with pattern preservation) hinder users from expressing intended behavior. This visualization illustrates how each misalignment type is represented in the trajectory structure, providing insights for interface and model design.}
\Description{Sub-trajectories in ARC Task 53b68214, where the proposed Misalignment Detection Algorithm (Alg.~\ref{alg:misalignment_detection}) identifies types of user misalignment. Red-circled regions indicate \textbf{Cognitive Dissonance in Users}, typically involving cyclic behavior such as undoing actions or navigating without a clear goal. Yellow circles highlight \textbf{User Unfamiliarity with Tools}, where users applied redundant sequences (e.g., repeated paint operations) to achieve outcomes that simpler actions could accomplish. Green ovals mark \textbf{Functional Inadequacies in Tools}, where limitations (e.g., no resize with pattern preservation) hinder users from expressing intended behavior. This visualization illustrates how each misalignment type is represented in the trajectory structure, providing insights for interface and model design.}
\label{fig:compare_tasks_difficulties}
\end{figure}

We applied our proposed Misalignment Detection Algorithm (Alg.~\ref{alg:misalignment_detection}) to Task 53b68214 to show its ability to detect misalignment types in user trajectories. 
The algorithm analyzes trajectory structure, action types, and recurring patterns characteristic of the three misalignment types, then labels the relevant sub-trajectories accordingly.

First, we identified \textbf{Functional Inadequacies in Tools} when users showed valid intentions that could not be executed due to interface limitations or missing functionality.
In Task 53b68214, many attempted to resize the grid while maintaining the existing layout, reflecting a natural user expectation.
However, the tool resets the grid upon resizing, erasing prior content, and interrupting the workflow.
This required users to manually recreate progress from scratch, introducing friction into what was otherwise a straightforward plan.
These sub-trajectories, marked in \textbf{green}, appear early in the task and reflect a gap between user expectations and tool capabilities, especially for grid-preserving transformations.

Second, the algorithm highlights \textbf{User Unfamiliarity with Tools} by spotting unnecessarily long and repetitive sub-trajectories.
A typical case is using the \texttt{paint} action two or three times to color individual pixels, despite support for multi-pixel selection or area-filling.
These sequences, which a single action could easily replace, are marked in \textbf{yellow} and usually appear near the task’s end, reflecting limited tool fluency or interface knowledge.
This suggests that the user grasps the high-level goal but not the full capabilities of the available toolset, resulting in redundant effort.

Finally, we identified instances of \textbf{Cognitive Dissonance in Users} by detecting cycles in trajectories. 
For example, when users executed a \texttt{paste} followed by an \texttt{undo}, or returned to a prior state after exploratory steps. 
These patterns suggest disorientation or indecision, deviating from goal-directed reasoning and indicating uncertainty about the correct course of action. 
Such behavior often reflects a breakdown in planning or inconsistency in the user’s evolving mental model. 
These segments are marked in \textbf{red} in Fig.~\ref{fig:compare_tasks_difficulties}, often near the center of the graph where exploration is dense.

This structured misalignment detection provides targeted feedback for both interface and model design improvements.
Differentiating between user error and tool limitation enables more informed and impactful design decisions for developers.
Understanding user struggles supports UI refinement, curriculum guidance, and model alignment with authentic human reasoning patterns.
Mapping misalignments on the trajectory graph reveals their frequency, spatial distribution, and association with task complexity and interaction patterns, helping to identify recurring issues and guide future improvements to tools and datasets.

\end{document}